\documentclass{article}

\usepackage{arxiv}

\usepackage[utf8]{inputenc} 
\usepackage[T1]{fontenc}    
\usepackage{hyperref}       
\usepackage{url}            
\usepackage{booktabs}       
\usepackage{amsfonts}       
\usepackage{nicefrac}       
\usepackage{microtype}      
\usepackage{lipsum}
\usepackage{graphicx}
\usepackage{amsmath}
\usepackage{multirow, array}
\graphicspath{ {./images/} }
\usepackage[skip=2pt,font=small]{caption}
\usepackage{atbegshi}
\AtBeginDocument{\AtBeginShipoutNext{\AtBeginShipoutDiscard}}

\usepackage[numbers]{natbib}
\usepackage{lipsum}
\usepackage{subcaption}
\usepackage{enumitem}
\usepackage{siunitx}

\usepackage{framed} 
\usepackage{multicol} 
\usepackage{nomencl} 
\makenomenclature

\providecommand{\keywords}[1]
{
  \small	
  \textbf{\textit{Keywords---}} #1
}

\title{A comparative assessment of deep learning models for day-ahead load forecasting: Investigating key accuracy drivers}

\author{
 Sotiris Pelekis \\
  Decision Support Systems Laboratory
  School of Electrical and Computer Engineering\\
  National Technical University of Athens\\
  Greece\\
  \texttt{spelekis@epu.ntua.gr} \\
   \And
 Ioannis-Konstantinos Seisopoulos \\
  Decision Support Systems Laboratory
  School of Electrical and Computer Engineering\\
  National Technical University of Athens\\
  Greece\\
  \texttt{giannissiso7@gmail.com} \\
  \And
 Evangelos Spiliotis \\
  Forecasting and Strategy Unit
  School of Electrical and Computer Engineering\\
  National Technical University of Athens\\
  Greece\\
  \texttt{spiliotis@fsu.gr} \\
  \And
Theodosios Pountridis \\
  Decision Support Systems Laboratory
  School of Electrical and Computer Engineering\\
  National Technical University of Athens\\
  Greece\\
  \texttt{theopountridis@gmail.com} \\
  \And
Evangelos Karakolis \\
  Decision Support Systems Laboratory
  School of Electrical and Computer Engineering\\
  National Technical University of Athens\\
  Greece\\
  \texttt{vkarakolis@epu.ntua.gr} \\
  \And
Spiros Mouzakitis \\
  Decision Support Systems Laboratory
  School of Electrical and Computer Engineering\\
  National Technical University of Athens\\
  Greece\\
  \texttt{smouzakitis@epu.ntua.gr} \\
  \And
Dimitris Askounis \\
  Decision Support Systems Laboratory
  School of Electrical and Computer Engineering\\
  National Technical University of Athens\\
  Greece\\
  \texttt{askous@epu.ntua.gr} \\
}

\begin{document}
\maketitle
\begin{abstract}
Short-term load forecasting (STLF) is vital for the effective and economic operation of power grids and energy markets. However, the non-linearity and non-stationarity of electricity demand as well as its dependency on various external factors renders STLF a challenging task. To that end, several deep learning models have been proposed in the literature for STLF, reporting promising results. In order to evaluate the accuracy of said models in day-ahead forecasting settings, in this paper we focus on the national net aggregated STLF of Portugal and conduct a comparative study considering a set of indicative, well-established deep autoregressive models, namely multi-layer perceptrons (MLP), long short-term memory networks (LSTM), neural basis expansion coefficient analysis (N-BEATS), temporal convolutional networks (TCN), and temporal fusion transformers (TFT). Moreover, we identify factors that significantly affect the demand and investigate their impact on the accuracy of each model. Our results suggest that N-BEATS consistently outperforms the rest of the examined models. MLP follows, providing further evidence towards the use of feed-forward networks over relatively more sophisticated architectures. Finally, certain calendar and weather features like the hour of the day and the temperature are identified as key accuracy drivers, providing insights regarding the forecasting approach that should be used per case.
\end{abstract}




\maketitle
\begin{table*}[!t]   
\begin{framed}
\nomenclature{CNN}{Convolutional neural network}
\nomenclature{DL}{Deep learning}
\nomenclature{EPES}{Electrical power and energy system}
\nomenclature{LSTM}{Long short-term memory}
\nomenclature{$l$}{Look-back window}
\nomenclature{MAPE}{Mean absolute percentage error}
\nomenclature{MLOps}{Machine learning operations}
\nomenclature{MLP}{Multi-layer perceptron}
\nomenclature{MLR}{Multiple linear regression}
\nomenclature{N-BEATS}{Neural basis expansion analysis time series forecasting}
\nomenclature{NN}{Neural network}
\nomenclature{RNN}{Recurrent neural network}
\nomenclature{RMSE}{Root mean squared error}
\nomenclature{STLF}{Short-term load forecasting}
\nomenclature{sNaive}{Seasonal naive}
\nomenclature{TFT}{Temporal fusion transformer}
\nomenclature{TCN}{Temporal convolutional network}
\nomenclature{TPE}{Tree-structured parzen estimator}
\printnomenclature
\end{framed}
\end{table*}

\section{Introduction}\label{sec:introduction}
\subsection{Background}
Electricity load forecasting is crucial for the optimal operation and modernization of power systems (e.g. smart grids) and has always been an important task for all countries worldwide irrespective of their economical state. This is because power systems are the backbone of modern life, enabling the technology that organizations and individuals possess to function within society. Load forecasts contribute in preserving the balance between electricity consumption and production, the economic power dispatch, storage scheduling, network planning, and expansion of power grids. However, electricity demand depends on numerous external variables \citep{Spiliotis2021a}, such as energy prices, weather conditions and demographics \citep{Psiloglou2009FactorsAssessment}, as well as social and political factors like the climate crisis and the recent outbreak of the Russian-Ukrainian war that urge for an immediate energy transition through an even higher penetration of RES \citep{Steffen2022APolicies}. These factors complicate the patterns of load time series and therefore sophisticated models are required to accurately extrapolate them in time.

From a technical perspective, load forecasting is typically divided into four categories according to the forecasting horizon considered. Long-term load forecasting involves forecasts up to 20 years ahead and is usually linked to grid development and strategic planning. Medium-term load forecasting involves a week up to a year ahead forecasts and is mainly used for maintenance scheduling and fuel purchases planning, as well as for energy trading and revenue assessment. Short-term load forecasting (STLF) includes one day to one week ahead forecasts. The resolution of said forecasts ranges from 15-minutes to one hour and they usually serve the day-to-day operations of utilities and companies participating in electricity markets. Lastly, very short-term load forecasting addresses forecast horizons of few minutes to few hours, being mostly used for demand response and near real-time control \citep{Hammad2020MethodsReview}.

With respect to STLF, it is considered a challenging task as electricity load time series are characterized by non-linearity, non-stationarity, noise, and multiple seasonal patterns. This can be attributed to the fact that electrical power demand originates from various electrical loads that, in turn, depend on numerous external variables, including weather and calendar factors. Additionally, the increased diversity of the behavior of the users connected to smart devices and electric vehicles has increased the fluctuation of the electricity load \citep{Nalcaci2019Long-termMethods}. Therefore, given the necessity for higher penetration of RES, more accurate forecasts are required that will effectively adapt to the flexibility and demand response requirements posed within transmission and distribution operation systems \citep{Pelekis2023TargetedTechniques}. As a result, STLF has been investigated in a variety of research studies and projects \citep{Bahrami2021DeepNetworks, Pelekis2022InPerformance, Karakolis2022ArtificialProject, Pau2022MATRYCSDomain, Wehrmeister2022TheApplications}. This research has led to an abundance of related literature that promotes continuous innovation and evolution in the field. Nonetheless, it has also created a fuzzy landscape for industrial electrical power and energy system (EPES) stakeholders that seek to identify the most suitable methods for their STLF use cases with a minimal effort.

The role of accurate STLF is crucial within power systems. According to \citet{Hobbs1999AnalysisForecasts}, a decrease of the mean absolute percentage error (MAPE) by 1\% can reduce generation costs by approximately 0.1\%-0.3\% when MAPE is in the range of 3\%-5\%. In this context, a conservative estimate is that a 1\% reduction in forecast error for a 10 GW utility can save up to \$ 1.6 million annually. \citet{Ortega-Vazquez2006EconomicInterruptions} explored the economical impact of load forecast errors on the daily power system operation considering the probability of generating units outages to estimate the energy not served due to the outages, concluding that an increase of 1\% in the error can lead to an economic impact of 590.14\$ daily, assuming an one-area IEEE-RTS system of 3105 MW. Meanwhile, \citet{StimmelCarol2019BigGrid} estimated that a 0.1\% improvement in forecasting for a midsize European utility could lead to a reduction of around \$3 million in operating costs within imbalance markets. This estimate took however into account the ability to forecast at a granular level, including meters and transformer feeders, thus enabling dynamic tariffs, demand response, and effective management of distributed energy resources that alleviate peak demand.

\subsection{Related work} \label{sec:1.2}

Recently, neural networks (NNs) and deep learning (DL) models have dominated the field of time series forecasting, including STLF among other energy forecasting applications. The multi-layer perceptron (MLP) architecture has been widely used for STLF \citep{Ho1992ShortAlgorithm, Kandil2006AnNetworks, Arvanitidis2021EnhancedNetworks}, even though not originally designed to model time series data that are chronologically correlated. Other approaches that include genetic algorithms have also been proposed to reduce training time \citep{Liao2006ApplicationForecasting, Chen2016Group-basedForecasting} and to select the training parameters of genetic algorithm neural network hybrid models \citep{VesaAndreeaValeria2020EnergyPrograms, Mishra2008ShortOptimization}. NNs have also been trained using artificial immune system models, mainly for hyperparameter tuning purposes \citep{Hernandez2013APlants, Farsi2021OnApproach}, and implemented in the form of extreme learning machines \citep{Zhang2013Short-termMachine}. Moreover, several ensembling techniques have been introduced to reduce model and parameter uncertainty and, consequently, improve forecast accuracy \citep{VesaAndreeaValeria2020EnergyPrograms, DeFeliceMatteo2011Short-TermNotes}. An exhaustive review on NN methods for STLF has been conducted by \citet{Hernandez2014ABuildings}. 

With respect to more recent advances, a massive flow of research studies have shifted towards deeper architectures  \citep{Gasparin2022DeepCase}. Recurrent neural networks (RNN) and particularly long short-term memory (LSTM) neural networks \citep{Hochreiter1997a} and their variants, such as sequence-to-sequence models and encoder-decoder architectures, cover a significant share of the current research interest on STLF \citep{Sehovac2020DeepAttention, Rueda2021Short-termGrid, Henselmeyer2021Short-termTraining}. Other innovative approaches build on convolutional neural networks (CNN) \citep{OShea2015AnNetworks} and LSTM hybrids since they can efficiently correlate load data with numerous weather variables \citep{Sajjad2020AForecasting, Rafi2021ANetwork}. 

Cutting-edge DL architectures include temporal convolutional networks (TCN) \citep{Bai2018AnModeling}, neural basis expansion coefficient analysis (N-BEATS) \citep{Oreshkin2019N-BEATS:Forecasting}, and transformers \citep{Lin2022ATransformers}. These models have recently drawn considerable attention in the field of STLF as they can result to relatively higher accuracy, lower training times, and more interpretable forecasts \citep{Lim2021TemporalForecasting, Oreshkin2019N-BEATS:Forecasting}. Specifically, \citet{Tang2022Short-TermNetwork} employed a TCN architecture with channel and temporal attention mechanisms to exploit the non-linear relationships between weather factors and load. \citet{Yin2021Multi-temporal-spatial-scaleSystems} applied a multi-temporal-spatial-scale TCN for forecasting the load of a city in China, while \citet{Gu2020TemporalModel} compared the TCN architecture with three traditional models, suggesting its superiority when used to forecast the load of a certain region in Shanghai. Regarding N-BEATS, it has been evaluated by \citet{Oreshkin2021a} on medium-term load forecasting tasks, achieving state-of-the-art performance, and by \citet{Singh2022Short-TermN-BEATS}, who employed the model for STLF in Ontario. \citet{Wen2021ProbabilisticIntervals} and \citet{Grabner2022APreprint} have also used variants of N-BEATS for probabilistic STLF and consumer-level STLF at global scale, respectively. In addition, \citet{Pelekis2022InPerformance} compared the accuracy of TCN and N-BEATS in STLF settings for the case of the Portuguese national load at a 15-minute resolution, with N-BEATS resulting to superior forecasts. With respect to transformers, \citet{Zhang2022Short-TermTransformer} evaluated a time augmented transformer for STLF on the electrical load data of New South Wales in Australia, \citet{Lim2021TemporalForecasting} validated temporal fusion transformers (TFT) for STLF on the UCI Electricity Load Diagrams Dataset \citep{UCI2022UCISet}, while \citet{Huy2022Short-TermModel} employed the latter architecture to forecast the load of Hanoi city using weather and calendar features.

Another research topic of great interest and implications in time series forecasting is the explainability of forecasting performance. \citet{Spiliotis2020AreReality} investigated the time series features of popular forecasting competition data sets, concluding that different forecasting methods should be used depending on the particularities of the data. Their analysis built on established feature extraction and visualization techniques \citep{Kang2017VisualisingSpaces}, as well as a multiple linear regression (MLR) model that correlated time series feature values with forecast accuracy. This type of explanatory analysis was first introduced by \citet{Petropoulos2014HorsesForecasting}, proposing the selection and combination of different forecasting models based on their expected forecast accuracy, predicted based on key time series features. In a similar direction,  \citet{Montero-Manso2020FFORMA:Averaging} used a tree-based model to combine the forecasts of various models based on time series features, while \citet{Talagala2022FFORMPP:Prediction} introduced a meta-learning algorithm for time series forecast model performance prediction and selection, also providing useful insights on which forecasting models best work for particular types of time series. In the STLF domain, \citet{Moon2022TowardValues} proposed an explainable tree-based model to forecast the demand in buildings using Shapley values \citep{Rozemberczki2022TheLearning}, concluding that the temperature-humidity index and the wind chill indexes affect the forecasting accuracy more than temperature, humidity, and wind speed. Similarly, \citet{Wu2022AnLearning} used Shapley values to develop a feature selection strategy for load forecasting models within a regional integrated energy system.


\subsection{Contribution} \label{sec:1.3}
It becomes evident that accurate STLF is invaluable for EPES stakeholders, while the options available in terms of modeling abundant. In this context, it is critical to evaluate the relative accuracy of well-established STLF models and, more importantly, to investigate the conditions under which each model is expected to perform best. The contribution of this study builds on such a comparative assessment, considering indicative DL architectures and key external factors (calendar and weather features) to explore the day-ahead forecasting accuracy of state-of-the-art forecasting approaches. Our innovations are summarized as follows:

\begin{enumerate}[label=(\alph*)]
    \item Considering the prevalence of RNNs and specifically LSTM as the golden standard in STLF applications, in this study we attempt to assess its accuracy over alternative DL architectures. In this context, we consider five DL models--some of them well-established, while others more recently introduced--and evaluate their accuracy in day-ahead load forecasting. The models include networks from the MLP (traditional feed-forward), LSTM (recurrent), TFT (hybrid recurrent/ feed-forward), TCN (convolutional), and N-BEATS (deep feed-forward with residual stacking) architectures that have recently drawn significant attention in the field of STLF. Considering univariate forecasting setups, we eventually confirm that feed-forward architectures, such as N-BEATS and MLP, can provide superior results despite the streamlined usage of RNNs. To ensure the representativeness of our results, we carefully tune the hyperparameter values of the models and produce forecasts for a complete calendar year using ensembles. Moreover, we benchmark the performance of the DL models using standard baseline models.
   
    \item Considering the importance of model performance explanation and selection based on key external variables, as well as the particularities of each data set that may affect forecast accuracy and favor certain models, we use MLR to correlate calendar (time of day, season, holidays, and weekends) and weather (temperature) features with the forecast errors of each DL model. Effectively, this post-hoc analysis enables us to decompose the forecasting performance of the examined models and identify the factors that deteriorate their accuracy most. Moreover, it allows us to identify models that are expected to perform best based on the calendar and weather features of the period being forecast. This analysis results in a set of guidelines for EPES stakeholders, such as transmission system operators, that are interested in reinforcing their forecasting toolset with state-of-the-art DL techniques or in generating forecasts in a selective fashion based on the state of external variables. To the best of our knowledge, this is the first attempt to apply such a model explanation approach--originating from the domain of operational research \citep{Petropoulos2014HorsesForecasting, Kang2017VisualisingSpaces, Spiliotis2020AreReality}-- to an STLF context.
    
    
    \item We expand the analysis conducted by \citet{Pelekis2022InPerformance} in three dimensions. First, we focus on purely autoregressive models, thus assessing the performance of DL time series forecasting models that do not require weather forecasts or any other type of external information as additional input. The reason for conducting this analysis is the ease of developing and deploying said models, alongside the absence of the requirement for high quality external and specifically weather data or forecasts that are typically unavailable for free. Nevertheless, we also compare the performance of purely autoregressive model to that of models that use weather forecasts as additional inputs, concluding that, most often that not, weather forecasts can indeed improve accuracy to some notable extent. Subsequently, we utilize high performance computing infrastructures to properly tune the examined models and evaluate their performance when used individually or within ensembles. By doing that, we investigate the robustness of each neural network architecture to random neural weight initializations and confirm the benefits of combining. Third, our analysis is performed on a newer version of the data set (ranging from 2013 to 2021 instead of 2019) that has a resolution of one hour instead of fifteen minutes.
    

\end{enumerate}

\subsection{Structure of the paper}

The rest of the paper is organized as follows. Section \ref{sec:2} describes our methodological approach, including the utilized DL architectures and benchmarks, the data used for the empirical evaluation, and the MLR framework employed to explain the forecasting performance of each model. Subsequently, Section \ref{sec:3} presents our results, while Section \ref{sec:4} discusses the impact of including the temperature in Lisbon as predictor within our DL models. Finally, Section \ref{sec:5} presents the key concluding remarks of our study, while Section \ref{sec:6} wraps up with potential future perspectives.

\section{Methodology}\label{sec:2}

This section presents the methodology used to address the common stages of the machine learning life-cycle referring to our specific STLF case in Portugal. These stages include the following tasks: (i) data collection, wrangling and transformations; (ii) exploratory analysis of the data set; (iii) selection and description of utilized DL architectures; (iv) training and model validation (hyperparameter tuning); (iv) forecast evaluation. Subsequently, the MLR-based framework used for explaining the forecasting performance of the DL models is presented. The experimental process took place using an automated machine learning operations (MLOps) pipeline developed with MLflow \citep{Alla2021}, building up to the one described by \citet{Pelekis2022InPerformance}.

\subsection{Data collection and exploratory analysis} \label{sec:2.1}

The data set used in the present study consists of the time series of the Portuguese national net aggregated electricity demand, reported at an hourly time resolution. The data was provided by R\&D Nester \citep{RDNester2023RDHomepage} in the context of the I-NERGY \citep{Karakolis2022ArtificialProject} project. The time series ranges from 2013 to 2021 as this was the most recent version of the data set including complete calendar years at the time the experiments were conducted. The operations applied to the electricity load time series before data scaling and model training took place were the following: (i) removal of duplicate entries (mainly caused by changes between standard and daylight saving time) and (ii) filling of missing data (rare cases handled using linear interpolation). An additional data source used was the historical weather archive in Lisbon that can be found online \citep{rp52023WeatherLisbon}. Specifically, hourly samples from the weather station "08579" were collected and an additional data set corresponding to the Portela airport from the same provider \citep{rp52023WeatherMETAR} was used in an auxiliary fashion to fill any missing values.

\begin{figure*}
\centering
\begin{subfigure}[b]{.45\linewidth}
   \centering
   \includegraphics[width=\textwidth]{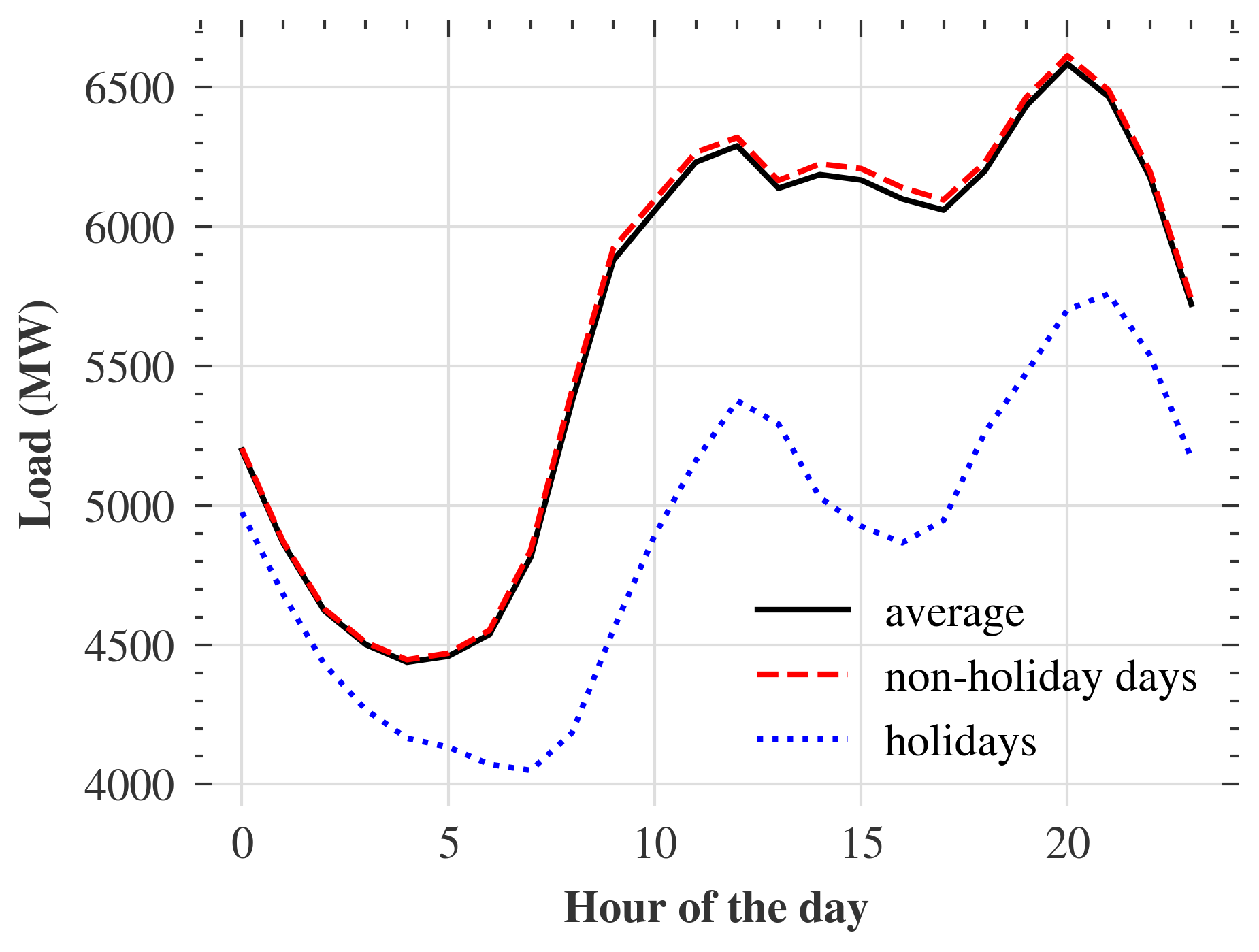}
   \caption{Average daily electricity load across holidays and non-holiday days.}
   \label{fig:1.1}
\end{subfigure}
\hfill
\begin{subfigure}[b]{.45\linewidth}
   \centering
   \includegraphics[width=\textwidth]{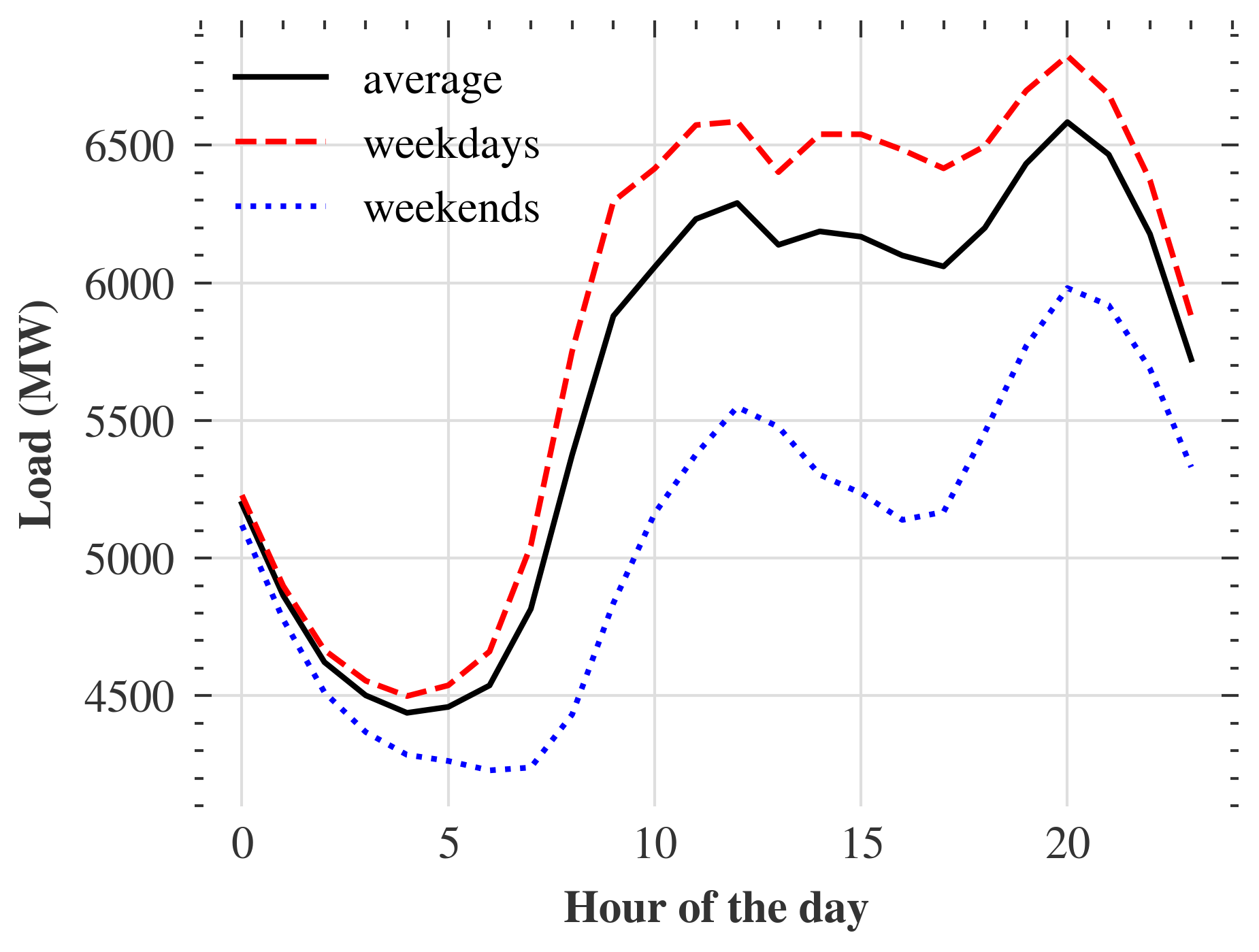}
   \caption{Average daily electricity load across weekends and weekdays.}
   \label{fig:1.2}
\end{subfigure}
\par\bigskip
\begin{subfigure}[b]{.45\linewidth}
   \centering
   \includegraphics[width=\textwidth]{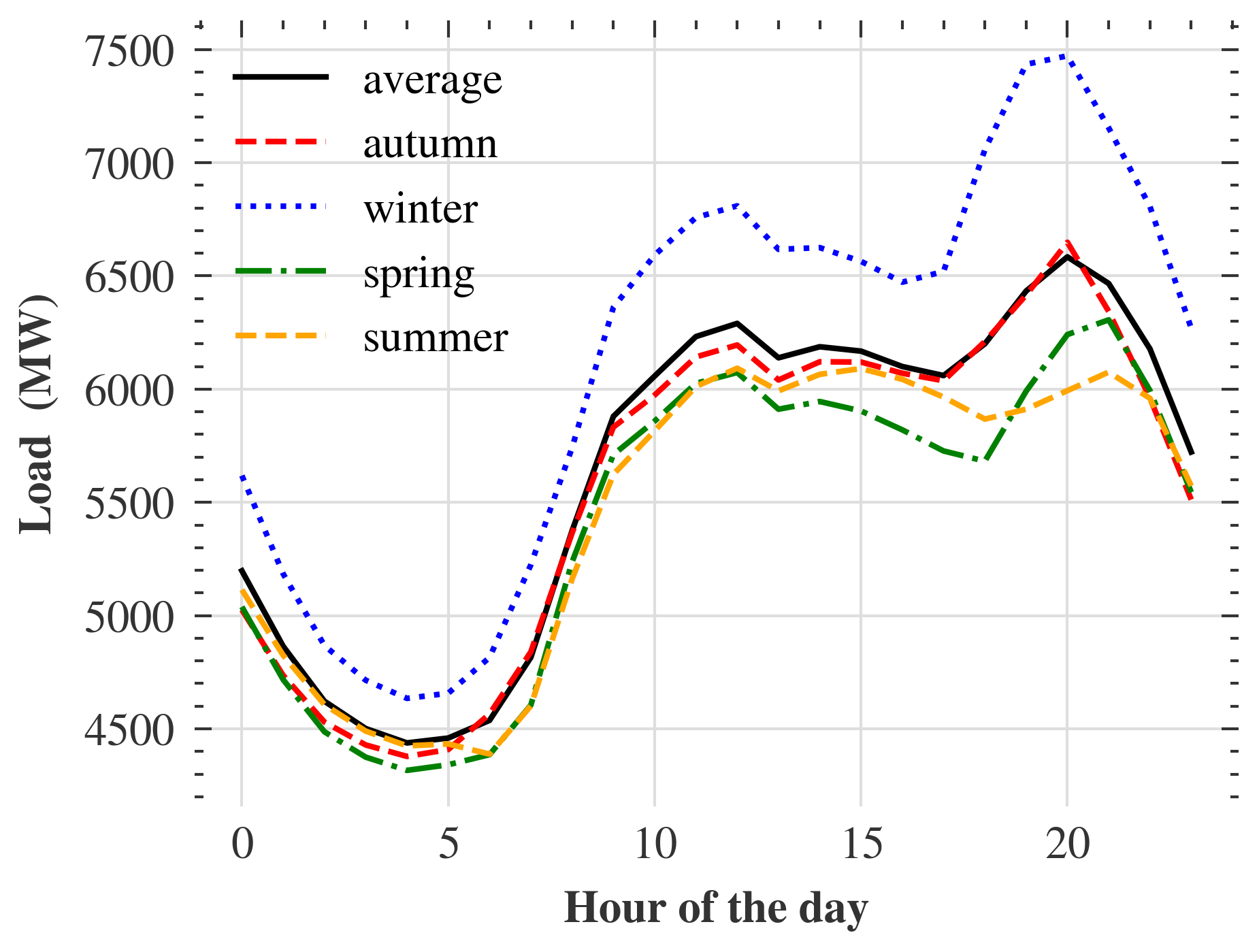}
   \caption{Average daily electricity load across the four different seasons of the year.}
   \label{fig:1.3}
\end{subfigure} 
\hfill
\begin{subfigure}[b]{.45\linewidth}
   \centering
   \includegraphics[width=\textwidth]{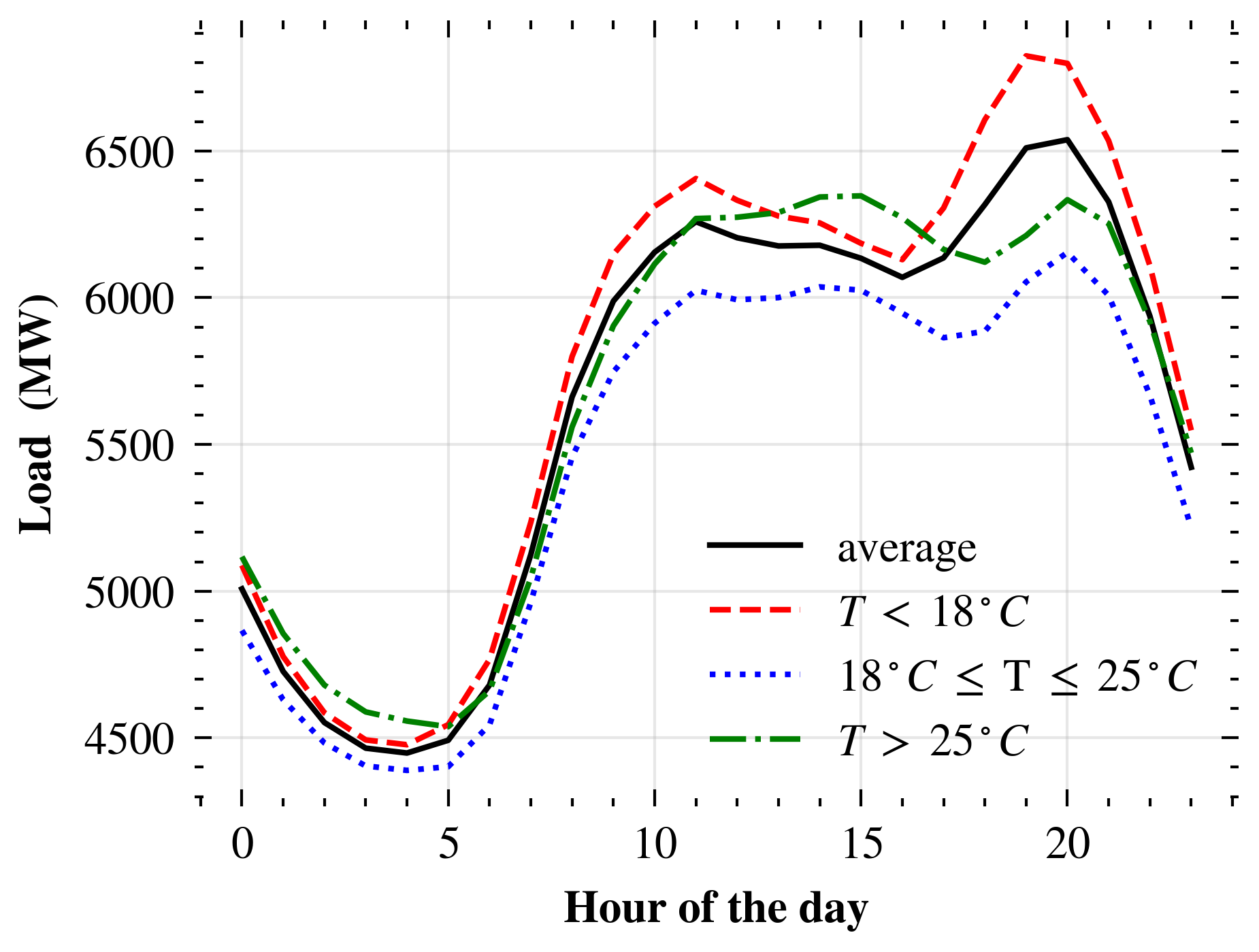}
   \caption{Average daily electricity load across different average daily temperature levels.}
   \label{fig:1.4}
\end{subfigure}

\caption{Average daily profiles of the Portuguese national electricity load time series grouped by selected calendar features(season of the year, holidays weekends) and weather features (temperature in Lisbon). The average load curve is also included in these graphs as a point of reference}
\label{fig:1}
\end{figure*}

Autoregressive models can be used to predict the future values of a time series by looking back at a predefined number of its past values. However, as already discussed, various exogenous variables can affect electricity demand, thus influencing the performance of forecasting models. To demonstrate this effect, we analyze the daily load profiles of the time series in juxtaposition to these exogenous factors. Figure \ref{fig:1} demonstrates the variation of daily load profiles given certain calendar features. All figures from  \ref{fig:1.1} to \ref{fig:1.4} illustrate the average daily load of Portugal as a reference point, while they additionally respectively depict: i) the average daily load on holidays versus non-holiday days, ii) the average daily load on weekends versus weekdays, ii) the inter-season daily profiles and ultimately iv) the daily profiles across different average daily temperature levels. 

Regarding the intra-day fluctuations, derived the average daily profile (black curve), four main intervals of varying time series behavior can be extracted:
\begin{itemize}
    \item \textbf{early morning} (23.00-04.00): During these hours an almost linear and decreasing pattern can be observed in the load profile given that most companies do not operate and households gradually reduce their electrical energy consumption.
    \item \textbf{morning} (04.00-09.00): A steadily increasing trend is observed within the load profile curve as every day life gradually recovers to its usual patterns;
    \item \textbf{midday} (09.00-17.00): This is a volatile period of the day corresponding to ordinary working hours. The load curve is noisy, however it maintains a rather constant level within this interval.
    \item \textbf{night} (17.00-23.00): A convex curve can be observed, corresponding to the consumption of peak hours when people get back from work, followed by the late night drop caused by the termination of most daily activities. 
\end{itemize}

\begin{figure}[]
\centering
   \includegraphics[width=0.6\linewidth]{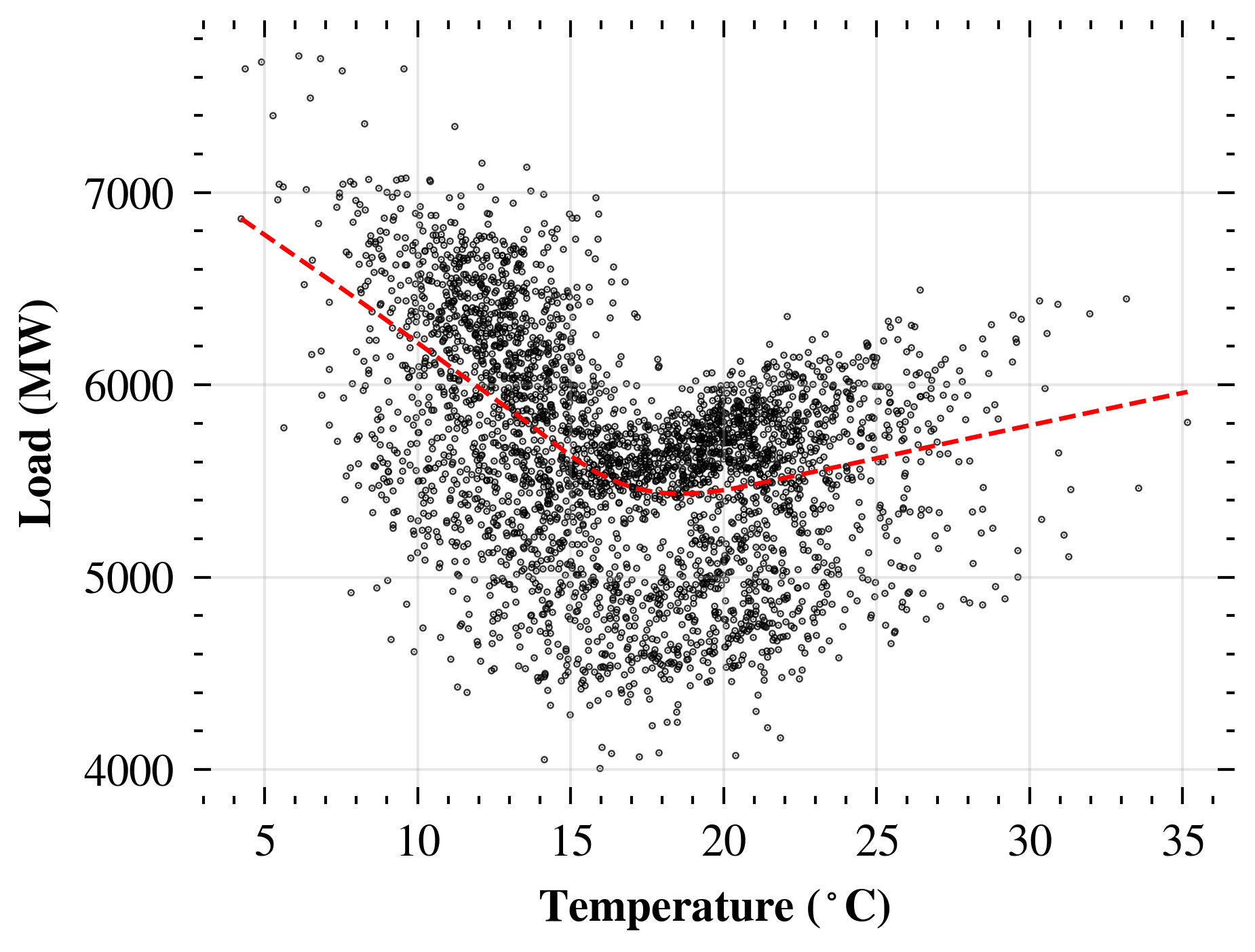}
   \caption{Scatter plot of the average daily electricity load across the respective average daily temperature levels. The locally weighted regression \citep{Cleveland1979RobustScatterplots} line is also included in the graph with red color.}
\label{fig:x}
\end{figure}

Considering the external variables, we observe that on holidays: a) the lower load values during the early morning hours are shifted towards later hours of the day as citizens tend to wake up later; b) the load exhibits a lower mean value than on non-holiday days; c) a notable load decrease during working hours; d) a peak shift towards later hours regarding the usual night peak of 8 p.m. Similar conclusions can be drawn for weekends and weekdays with the exception of the late night peak shift. Regarding the per season average load profiles, the deviations among seasons are still visible with the naked eye. During the winter period a significant increase of the average load and peaks is observed due to the presence of heating loads alongside the full operation of industrial/working activities. Autumn profiles exhibit a similar behavior to the average load profile, which is expected due to the presence of both hot and cold days during this season. Summer and spring profiles tend to exhibit lower load levels in general, suggesting that high load values are mainly driven by higher heating demand, especially at night hours when temperatures are typically lower. On the contrary, cooling loads seem to mainly affect late midday hours when the temperature is higher and this is exactly the interval when the summer profile is above that of spring. Note that summer is also linked with holidays and therefore the average level of the load profile is expected to be lower compared to other seasons as, except for cooling loads during the day, the rest of the loads are functioning to a smaller extent.

Nevertheless, as seasons involve mixed temperature levels and the effect of the latter cannot be directly observed, we additionally include Figure \ref{fig:1.4} that illustrates the effect of average daily temperatures on the average load curve. Note here that the thresholds for temperature levels have been chosen based on the the Portuguese regulation on the energy performance of residential buildings (REPRS) \citep{REPRS2013RegulationBuildings, RicardoAguiarRevisao2013ClimatologiaEdificios}. These temperatures are 18$^{\circ}$C  for heating degree-days (HDD) and 25$^{\circ}$C for cooling degree-days (CDD). Specifically, we observe that the shape of the daily load profile for average daily temperatures lower than 18$^{\circ}$C (cold days) is similar to the winter period, exhibiting however significantly lower magnitudes. This can be attributed to the fact that the average temperature of winter months within our data set is 12.08$^{\circ}$C, therefore implying that lower demand for heating loads is expected for average daily temperatures below 18$^{\circ}$C, which apparently do not correspond to pure winter days. Subsequently, for high temperature profiles we can also observe that the respective load curve exhibits high magnitudes, especially during the midday where usually the demand from cooling loads is increased while a certain decrease is notable during the night, despite the fact that people get back to home, as lower temperatures will be normally be observed there. To validate the assumption that hourly load values and daily load profiles are strongly dependent on the temperature levels, we also include the scatter plot of average daily loads against the respective average daily temperatures. We observe a bimodal joint distribution of said quantities leading to the conclusion that cold days can co-exist with extremely high as well as extremely low loads. The same can be observed for hot days. This behavior can be attributed to holidays and weekends (lower region) which, as observed from Figures \ref{fig:1.1} and \ref{fig:1.2}, exhibit on average much lower average loads contrary to working days (upper region). The locally weighted regression \citep{Cleveland1979RobustScatterplots} line (red color) that is also included in the graph clearly demonstrates the non-linear relationship among the two quantities, highlighting the fact that hot and especially cold days have a high impact on the increase of the load due to the functioning of cooling and heating systems, respectively.

The above observed intra-day patterns between the load and the external variables (calendar and temperature) reveal a certain correlation which implies potential performance drifts of different forecasting models across the different state of said variables. Hence, the motivation for explaining model errors  based on such external variables becomes indisputable and forms one of the main objectives of our study.

\subsection{Selected models and benchmarks} \label{sec:2.2}

To conduct our comparative analysis, an indicative set of models was considered to demonstrate the different aspects of current state-of-the-art DL architectures. When evaluating forecasting models, it is necessary to introduce concrete benchmarks. Therefore, two seasonal naive (sNaive) models of daily and weekly seasonality were first employed. Moreover, since our aim is to investigate the performance of the currently popular LSTM--which is selected as the representative of RNNs-- with that of alternative DL architectural setups, we also employ feed-forward NNs using MLP as a pure traditional NN and N-BEATS as a feed-forward approach with residual-stacking that is tailored for time series forecasting. Subsequently, we include TFT as a recurrent/feed-forward hybrid, and finally, to further extend the scope of our comparisons, we also employ TCN as a convolutional candidate. The selected DL models and benchmarks are presented below in more detail:
    
\paragraph{\textbf{sNaive (24)}} This naive model of daily seasonality uses the observed values of the previous day as day-ahead forecasts. The method is expected to generate poor results as electricity load time series exhibit multiple seasonal cycles, being strong at both daily and weekly level. 
    
\paragraph{\textbf{sNaive (168)}} This naive model of weekly seasonality uses the observed values of the same day of the previous week as day-ahead forecasts. The method is expected to produce moderate results as it can effectively take into account daily and weekly seasonality.

\paragraph{\textbf{Multi-layer perceptron (MLP)}} This is the most common DL architecture, consisting of multiple perceptrons (or neurons) arranged in multiple layers. Each individual neuron’s output is calculated using the function of Equation \ref{eq:00}.

\begin{equation}
y = f(\sum_{i=1}^{n} (w_ix_i) + b)
\label{eq:00}
\end{equation}
    
The weighted (with weights $w_i$) sum of the outputs ($x_i$) of all the neurons in the previous layer plus a bias term ($b$) is computed first. Then, the activation function $f$ is applied to this result. The purpose of this function is to enable the model to capture nonlinear relationships among the inputs (lags) and outputs (forecasts). Various activation functions have been used in literature, the most popular being the rectified linear unit (ReLu) of Equation \ref{eq:01}.
    
\begin{equation}
f(x) = \frac{1}{1 + e^{-x} }
\label{eq:01}
\end{equation}
    
During the training process of a MLP, the back-propagation algorithm is used. The weights and biases of the MLP are iteratively updated according to the following equation
    
    \begin{equation}
    w^{k}_{ij}(t+1) =  w^{k}_{ij}(t) - \alpha \frac{\partial L}{\partial w^{k}_{ij}},
    \label{eq:03}
    \end{equation}
    
\noindent where $w^{k}_{ij}(t)$ is the weight connecting the $i$-th neuron of a layer with the $j$-th neuron of the previous layer at iteration $t$ of the algorithm, $\alpha$ is the learning rate, and $L$ is the loss function. The gradient is computed using the chain rule, and the algorithm needs to be run after each forward pass of the training. 

In this study, the Scikit-Learn \citep{Pedregosa2011Scikit-learn:Python} machine learning toolkit, and specifically the MLPRegressor class, was used to implement the MLP model.
    
\paragraph{\textbf{Long short-term memory (LSTM)}} 
LSTM is an extension of the classical RNN that came to solve the vanishing gradient problem, mitigating RNNs' inability to learn long range dependencies \citep{Hochreiter1997a}. LSTM has been extensively used for sequence modeling and time series forecasting. It is capable of learning long-term dependencies by introducing a memory cell and three different gates, namely the input gate, output gate, and forget gate. This allows the network to store information over a large number of time steps, which is crucial for time series forecasting. The LSTM architecture can be represented mathematically as shown below:

\begin{equation}
\begin{aligned}
f_t &= \sigma(W_f \cdot [h_{t-1}, x_t] + b_f), \\
i_t &= \sigma(W_i \cdot [h_{t-1}, x_t] + b_i), \\
o_t &= \sigma(W_o \cdot [h_{t-1}, x_t] + b_o), \\
g_t &= \tanh(W_g \cdot [h_{t-1}, x_t] + b_g), \\
c_t &= f_t \cdot c_{t-1} + i_t \cdot g_t, \\
h_t &= o_t \cdot \tanh(c_t),
\end{aligned}
\label{eq:005}
\end{equation}

\noindent where $f_t, i_t, o_t$ are the forget, input, and output gates, $g_t$ is the cell state, $c_t$ is the memory cell, $h_t$ is the hidden state at time step $t$, and $b_k$ the respective bias of gate $k$. Backpropagation through time is used to train calculate the gradients of LSTM architecture and gradually update their weights.

Focusing on encoder-decoder architectures, they can be used for time series forecasting by recurrently encoding the input sequence into a fixed length vector, and then recurrently decoding it back into a sequence of predictions as the encoder and decoder are typically implemented using RNNs and more often LSTMs. The hidden dimension of the encoder and decoder is an important hyperparameter that needs to be the same for the two components in order for the model to function properly. Another important hyperparameter of an encoder-decoder LSTM is the number of recurrent layers in the encoder and decoder. Typically, increasing said number leads to more complex models that are prone to overfitting and generalize poorly on previously unseen data. Hence, the selection of these hyperparameters should be handled with care. The encoder-decoder architecture can be represented as shown below:

\begin{equation}
\begin{aligned}
h_t &= \text{LSTM}_{\text{encoder}}(x_t) \\
\hat{y}_t &= \text{LSTM}_{\text{decoder}}(h_t)
\end{aligned}
\label{eq:006}
\end{equation}

\noindent where $h_t$ is the hidden state of the encoder at time step $t$ and $\hat{y}_t$ is the forecast at the same step. 

In this study, the Darts \citep{Herzen2021Darts:Series} forecasting toolkit was used to implement the LSTM model, exploiting the BlockRNNModel class. 
    
\paragraph{\textbf{Neural basis expansion analysis time series forecasting (N-BEATS)}}
N-BEATS is a deep feed-forward neural network, introduced by \citet{Oreshkin2019N-BEATS:Forecasting} in an attempt to demonstrate that a pure autoregressive DL models can outperform traditional forecasting methods. The N-BEATS architecture consists of $n$ stacks, which comprise $m$ fully connected non-linear neural regressor blocks, interconnected with double residual links. Each block has a fully connected stack of $k$ layers using the ReLU activation function. The first block receives an input \textbf{$\hat{x}_1 \equiv \hat{x}$}, whereas the other blocks receive an input \textbf{$\hat{x}_m$}, producing two distinct outputs, namely the backcast \textbf{$\hat{b}_m$} and the forecast \textbf{$\hat{f}_m$}. The former (\textbf{$\hat{b}_m$}) is subtracted from the input of the next block as in Equation \ref{eq:000} and is referred to the part of the input that has been predicted adequately, while the latter (\textbf{$\hat{f}_m$}) is a partial forecast to predict the future.   

\begin{equation}
\hat{x}_m= \hat{x}_{m-1}-\hat{b}_{m-1}
\label{eq:000}
\end{equation}

This strategy is called double residual stacking and contributes to both forecasting and decomposing the target time series. The outputs of the blocks (\textbf{$\hat{f}_m$}) are aggregated to produce the forecast (\textbf{$\hat{f}_n$}) of each stack, as shown in Equation \ref{eq:001}, and it is used to calculate the loss function and train the model.

\begin{equation}
\hat{f}_n= \sum_{i=1}^{m} \hat{f}_m 
\label{eq:001}
\end{equation}

Finally, the outputs (\textbf{$\hat{f}_n$}) of the stacks are combined to produce the forecast (\textbf{$\hat{y}$}) of the model, as shown in Equation \ref{eq:002}.

\begin{equation}
\hat{y}= \sum_{i=1}^{n} \hat{f}_n 
\label{eq:002}
\end{equation}

In this study, the Darts forecasting toolkit was used to implement the N-BEATS model. The generic form of the architecture was used without considering the ensembles originally proposed by \citet{Oreshkin2019N-BEATS:Forecasting}.
    
\paragraph{\textbf{Temporal convolutional network (TCN)}} TCN, first introduced by \citet{Bai2018AnModeling}, is a type of neural network architecture that has been specifically designed for processing sequential data. Specifically, TCN is a CNN that consists of dilated, causal 1-dimensional convolutional layers with the same input and output lengths. Stack dilated convolutional networks are constructed to capture long term temporal dependencies, enabling large receptive fields with the use of a reasonable amount of layers. Equation \ref{eq:010} shows the basic formula characterizing a temporal convolutional layer

\begin{equation}
y[i] = b[i] + \sum_{j} w[i,j] * x[j],
\label{eq:010}
\end{equation}

\noindent where $x$ is the input sequence, $w$ is the weight matrix of the layer, $b$ is the bias vector, $y$ is the output sequence, while $i$ and $j$ are the indexes of the time step in the input and output sequence, respectively. The depth of the network is determined by the number of convolutional layers while the width of the network is determined by the number of filters in each convolutional layer. The size of the kernel plays a crucial role, allowing the network to learn more global patterns in the data. The residual block has a doublet of dilated convolutional layers followed by a batch normalization operation and a ReLU activation. Regarding the receptive field ($R$) of a TCN, it is usually calculated using the number of convolutional layers ($l$), the dilation base ($b$), and the kernel size ($k$), as shown in Equation \ref{eq:011}. The purpose of such calculation is to assure that the receptive field is large enough to process the input variables.

\begin{equation}
R = b^l \cdot (k-1) 
\label{eq:011}
\end{equation}


In this study, the Darts forecasting toolkit was used to implement the TCN model.

\paragraph{\textbf{Temporal fusion transformer (TFT)}}
The TFT is a powerful neural architecture for time series forecasting that combines the strengths of transformers \citep{Vaswani2017AttentionNeed} and autoregressive models. They were first introduced by \citet{Lim2019TemporalForecasting} and have been specifically designed to handle the challenges of temporal data, making them well-suited for accurate predictions in various forecasting tasks. TFT operate on the concept of encoding both the input features and the temporal context of the time series data. This is achieved through the integration of self-attention mechanisms within the transformer architecture. The self-attention mechanism enables TFT to capture the dependencies between different time steps, effectively modeling the temporal dynamics present in the data. The key component of a TFT is the fusion layer, which combines the encoded input features and the temporal context representations. The fusion layer utilizes gating mechanisms to adaptively weight the importance of each representation. This enables the model to effectively integrate both local and global temporal information, resulting in improved forecast accuracy. Mathematically, the fusion layer can be represented as in Eq. \ref{eq:x}.

\begin{equation}
\begin{aligned}
    \mathbf{z}_t = \sigma(\mathbf{W}_z[\mathbf{x}_t, \mathbf{c}_t] + \mathbf{b}_z), \\
    \mathbf{g}_t = \text{softmax}(\mathbf{W}_g[\mathbf{x}_t, \mathbf{c}_t] + \mathbf{b}_g), \\
    \mathbf{c}_t' = \mathbf{g}_t \odot \mathbf{c}_t + (1 - \mathbf{g}_t) \odot \mathbf{z}_t,
\end{aligned}
\label{eq:x}
\end{equation}
where $\mathbf{x}_t$ represents the input features at time step $t$, $\mathbf{c}_t$ is the temporal context representation at time step $t$, $\mathbf{W}_z$ and $\mathbf{W}_g$ are weight matrices, $\mathbf{b}_z$ and $\mathbf{b}_g$ are bias vectors, $\sigma$ denotes the activation function, $\text{softmax}$ is the softmax function, and $\odot$ represents element-wise multiplication. The fusion layer is typically applied iteratively across multiple time steps, allowing the TFT to progressively update the temporal context representation based on the input features. This iterative process enhances the model's ability to capture complex temporal patterns and improve forecast accuracy.

In this study, the Darts forecasting toolkit was used to implement the TFT model.

\subsection{Model training and validation} \label{sec:2.3}

The objective of the training process has been the development of DL models for day-ahead load forecasting at hourly resolution for the net aggregated electricity demand of the Portuguese transmission system. Regarding the model training pipeline, every DL model was: (i) trained on the train set (years 2013 to 2019: 61,344 data points); (ii) optimized on the validation set (year 2020: 8,784 data points) to identify appropriate hyperparameter values; (iii) and evaluated on the remaining, previously unseen test set (year 2021: 8,760 data points) without retraining the model on the full data set. This split allowed the inclusion of complete calendar years within each data set, thus maintaining the symmetry of seasonal patterns. Regarding the sampling procedure, the training set comprises $N_{train}$ samples as shown in Equation \ref{eq:200}.

\begin{equation}
N_{train} = k_{train} - l - f + 1 = 61,321 - l
\label{eq:200}
\end{equation}

\noindent where $k_{train}$ is the number of data points in the train set, $f$ the forecast horizon (24 hours), and $l$ the look-back window used by the model. Similarly, the validation set consists of $N_{val}$ samples, as shown in Equation \ref{eq:201}. 

\begin{equation}
N_{val} = k_{val} - f + 1 = 8,761
\label{eq:201}
\end{equation}

\noindent where $k_{val}$ is the number of validation data points. With respect to the testing process, the test set contains $N_{test}$ samples, as shown in Equation \ref{eq:202}, given that the model evaluation took place at a daily basis rather than hourly, considering daily forecast batches of 24 hours.
\begin{equation}
N_{test} = \frac{k_{test}}{f} = 365
\label{eq:202}
\end{equation}

Note here that the optimizers of each individual model have been programmed to  minimize the L2 loss, namely squared error loss, following a mini-batch gradient computation approach while considering all possible rolling windows i.e 00:00 - 23:00, 01:00 - 00:00 (next day), 02:00 - 01:00 (next day) etc. To prevent overfitting early stopping was applied with a patience of 10 epochs on the loss of the validation set. All models have been trained on the scaled version of the time series using a normalization approach within the [0,1] space. Note also that scaling took place using the range of values observed in the training set.

With respect to hyperparameter tuning, the selected DL architectures were trained on multiple sets of hyperparameter values using the tree-structured parzen estimator (TPE) hyperparameter optimization method, as described by \citet{Bergstra2011AlgorithmsOptimization} and implemented in Python programming language in Optuna optimization library \citep{Akiba2019Optuna:Framework}. A crucial hyperparameter that is common among all models, is the look-back window ($l$), namely the number of historical time series values (lags) that the model looks back at when being trained to generate forecasts. Note that $l$ directly determines the input layer size of the trained neural network. The specific details of the optimization process for each architecture are presented in Table \ref{tab:1} where only the hyperparameters that were tuned are presented, while the rest of the hyperparameters were set to their default values, as proposed by the respective Python implementation frameworks (Darts, Scikit-Learn), apart from some small exceptions that are listed in section \ref{sec:3.1}. A total of 100 trials were executed for selecting the optimal DL architecture based on the minimum MAPE value on the validation set. This time, the calculation of said metric was performed iteratively using a stride of 24 hours, hence selecting the optimal forecasting model for complete calendar days (only forecasts of the 00.00-23:00 intervals).

\begin{table}[]
\centering
\caption{The hyperparameters optimized per DL architecture and the respective search spaces.}\label{tab:1}
\begin{tabular}{@{}lll@{}}
\toprule
\textbf{Architecture} & \multicolumn{2}{l}{\textbf{Hyperparameters}} \\
~ & Name & Space \\  
\midrule
\multirow{5}{*}{MLP} & $l$ & \{24, 48,..., 480\} \\ 
~ & \# layers & \{1, 2,..., 10\}\ \\
~ & \# neurons per layer & \{32, 64, 128, 256, 512\} \\ 
~ & activation function & \{relu, sigmoid\}\\ 
~ & batch size & \{256, 512,..., 2048\}\\ 
\hline
\multirow{4}{*}{LSTM} & $l$ & \{24, 48,..., 480\}\\ 
~ & \# RNN layers & \{1, 2,..., 10\}\\ 
~ & hidden dimension size & \{24, 48,..., 240\} \\ 
~ & batch size & \{256, 512,..., 2048\}\\ 
\hline
\multirow{5}{*}{N-BEATS} & $l$ & \{24, 48,..., 480\}\\
~ & \# stacks & \{1, 2,..., 25\}\\ 
~ & \# blocks & \{1, 2,..., 20\}\\ 
~ & \# layers & \{1, 2,..., 10\}\\ 
~ & batch size & \{256, 512,..., 2048\}\\ 
\hline
\multirow{5}{*}{TCN} & $l$ & \{24, 48,..., 480\} \\ 
~ & kernel size & \{2, 3,..., 6\} \\ 
~ & \# filters & \{2, 3,..., 12\} \\ 
~ & dilation base & \{2, 4, 8, 16, 32\} \\ 
~ & batch size & \{256, 512,..., 2048\} \\ 
\hline
\multirow{5}{*}{TFT} & $l$ & \{24, 48,..., 480\} \\ 
~ & \# LSTM layers & \{1, 2, 3, 4\} \\ 
~ &  \# attention heads& \{1, 4\} \\ 
~ &  dropout & \{0, 0.1, 0.3, 0.5, 0.7, 0.9\} \\ 
~ &  hidden dimension size & \{16, 32, 64, 128, 256\} \\ 
~ & batch size & \{256, 512,..., 2048\} \\ 
\bottomrule
\end{tabular}
\end{table}

Following the identification of the best model architecture in terms of hyperparameter values, 30 separate networks sharing the same hyperparameters but different pseudo-random initializations of neural weights were trained for each DL architecture. Then, an ensemble of these models was used to produce the final forecasts on the test set. To improve the robustness of the results, the median operator was used to aggregate the forecasts of the individual models \citep{Kourentzes2014NeuralForecasting}. Note that the sub-models of each ensemble were trained on the union of the training and validation sets for a predefined number of epochs as established from the hyperparameter optimization stage therefore not requiring early stopping.

The training, validation, and ensembling of the models was executed on an Ubuntu 22.04 virtual machine with an NVIDIA Tesla V100 GPU, 32 CPU cores, and 64GB of RAM.

\subsection{Evaluation of forecast accuracy} \label{sec:2.4}
With respect to the evaluation of forecast accuracy, various forecasting performance measures are common in the literature, such as the mean absolute error (MAE), the mean squared error (MSE), the mean absolute percentage error (MAPE), the root mean square error (RMSE), the symmetric mean absolute percentage error (sMAPE), and the mean absolute scaled error (MASE) \citet{Hyndman2006}. Within the present study, the MAPE is considered as the primary evaluation measure, reporting the average magnitude of the absolute percentage errors. It is widely used in practice because it allows for easy interpretation and comparison across different forecasting models. By considering the relative magnitude of the errors in terms of actual values, the MAPE helps in understanding and communicating the overall accuracy of the forecasts. Therefore, the MAPE is a widely accepted choice in STLF applications and leads to explainable evaluations. Specifically, we used the MAPE as the objective function during the hyperparameter optimization process and as the main criterion for evaluating the models and investigating their appropriateness for different calendar and weather conditions. MAPE is defined as shown in Equation \ref{eq:1}.

\begin{equation}
MAPE=\frac{1}{m} \sum_{t=1}^{m}\left|\frac{y_{t}-\hat{y}_t}{y_{t}}\right| \cdot 100(\%)
\label{eq:1}
\end{equation}

\noindent where $m$ represents the number of samples, while $y_t$ and $\hat{y}_t$ stand for the actual values and the forecasts at time $t$, respectively. 

Nevertheless, MAPE has some limitations. It can be sensitive to extreme values and it may produce infinite or undefined values when the actual values are zero. Additionally, the MAPE treats positive and negative errors unequally, putting heavier penalty on negative errors (overcasting). In order to enhance the representativeness of our results and better support our discussion, we also use the RMSE as a secondary evaluation metric. The RMSE provides a symmetric measure of the dispersion or spread of the forecast errors. It is widely used because it gives higher weight to larger errors, thus penalizing larger deviations between predicted and actual values more heavily. By taking the square root of the average squared errors, the RMSE is expressed in the same units as the original data set, allowing for easy interpretation and comparison. The RMSE is calculated by taking the square root of the mean of the squared differences between the predicted values ($\hat{y}$) and the actual values ($y$) in the data set. Mathematically, it can be represented as:

\begin{equation}
RMSE = \sqrt{\frac{1}{m}\sum_{t=1}^{n}(\hat{y}_t - y_t)^2}
\end{equation}
where $m$ represents the number of observations in the data set.

Note here that the evaluation of the models on the validation (hyperparameter tuning) and test sets (final performance) was performed iteratively using a stride of 24 hours as also mentioned in the previous section.

\subsection{Multiple linear regression for forecasting error explanation and model selection} \label{sec:2.5}

Although forecasting accuracy measures like MAPE are useful for identifying models that perform best on average, they provide no information about the conditions under which the forecast accuracy of each model is typically improved or deteriorated. In order to provide intuitive insights in this direction, for each architecture, a MLR model was estimated to correlate the MAPE values of the test set (dependent variable) with nine key forecast accuracy drivers (independent variables), as follows: 

\begin{itemize}
    \item Three binary (one-hot encoded) features (F1: morning, F2: midday, F3: night), representing the \textbf{time of day}. The "early morning" variable was dropped from the model to avoid undesirable multicollinearity effects.
    \item Three binary features (F4: winter, F5: spring, F6: autumn), representing the \textbf{season of the year}. The "summer" variable was dropped to avoid undesirable multicollinearity effects.
    \item One binary feature (F7: holiday), representing the \textbf{Portuguese national holidays}. The "non-holiday" variable was dropped to avoid undesirable multicollinearity effects.
    \item One binary feature (F8: weekend), representing \textbf{weekends}. The "weekday" variable was dropped to avoid undesirable multicollinearity effects.
    \item One numerical feature (F9: temperature) representing the \textbf{temperature} in the city of Lisbon. This variable was introduced as an extra dependent variable in the MLR problem, as it is known to affect electricity load patterns \citep{Moghaddas-Tafreshi2008a, Cui2015Short-TermModel,Haben2019a}.
\end{itemize}

Hence, each of the MLR models can be formulated as follows:
\begin{equation}
MAPE_k = a{F1}_k + b{F2}_k + ... + h{F3}_k + i{F9}_k
\label{eq:2}
\end{equation}

\noindent where $MAPE_k$ is the error generated for the $k_{th}$ data point of the selected test set (year 2021) and ${Fi}_k$ the value of variable $Fi$ at the same point. Before estimating the models, both dependent (MAPE) and independent (external features) variables were 1\% trimmed in order to mitigate the effect of extreme values and enhance the representativeness of the results. 

Note that each MLR model can effectively serve as a framework for decomposing and explaining forecasting performance, predicting forecast accuracy and, consequently, allowing to conditionally  select the most appropriate forecasting model based on the specific needs of the use case posed by an EPES stakeholder.  


\section{Results}\label{sec:3}
In this section, the results of the study are presented and discussed, addressing two main aspects as follows: 
\begin{itemize}
    \item The macroscopic comparison and ranking of the selected DL architectures in terms of forecast accuracy in the examined STLF task (Section \ref{sec:3.1}).
    \item The detailed results of the MLR models, accompanied by explanatory comments regarding each architecture's performance, as well as model selection insights based on the selected calendar and weather features (Section \ref{sec:3.2}).
\end{itemize}

\subsection{Forecasting accuracy}\label{sec:3.1}

Initially, the results of the training and hyperparameter tuning processes are summarized, leading to the final architectures and weights for each of the selected forecasting methods. The computational times for running the 100 TPE trials are also listed along with the times required for training the models of each ensemble as they are indicative of the resources needed in real life for their implementation. The results are as follows.

\paragraph {\textbf{sNaive (24):}} No optimization took place as there are no hyperparameters to tune. The method required 1 minute to produce forecasts for the complete test set.
\paragraph {\textbf{sNaive (168):}} No optimization took place as there are no hyperparameters to tune. The method required 1 minute to produce forecasts for the complete test set.
\paragraph {\textbf{MLP:}} The TPE optimization process of the MLP architecture ran for 199 minutes to converge to the following hyperparameter values: $l$: 192, \# layers: 2, \# neurons per layer: [256, 256], activation function: relu, batch size: 1024. Given these values, each model of the ensemble was trained for 2.12 minutes on average, running for a total 109 epochs. Note here that the alpha parameter of the MLPRegressor class was set to 0 to avoid any regularization. The method required 1 minutes to produce forecasts for the complete test set.
\paragraph {\textbf{LSTM:}} The TPE optimization process of the LSTM architecture took 1,664 minutes to converge to the following hyperparameter values: $l$: 240, \# RNN layers: 3, hidden dimension size: 48, batch size: 1,280. Given these values, each model of the ensemble was trained for 22.6 minutes on average, running for a total 180 epochs. The method required 1 minutes to produce forecasts for the complete test set.
\paragraph {\textbf{N-BEATS:}} The TPE optimization process of the N-BEATS architecture took 2589 minutes and resulted to the following set of hyperparameter values: $l$: 192, \# stacks: 3, \# blocks: 6, \# layers: 3, batch size: 1,536. Note here that the layer widths and the dimension of the expansion coefficient have been set to 64 and 5 respectively. Given these values, each model of the ensemble was trained for 13.1 minutes on average, running for a total 108 epochs. The method required 1 minutes to produce forecasts for the complete test set.
\paragraph {\textbf{TCN:}} The TPE optimization process of the TCN architecture took 3,698 minutes to converge to  the following set of hyperparameter values: $l$: 456, kernel size: 4, \# filters: 10, dilation base: 8, batch size: 256. Given these values, each model of the ensemble was trained for 55 minutes on average, running for a total 296 epochs. The method required 1 minutes to produce forecasts for the complete test set.
\paragraph {\textbf{TFT:}} The TPE optimization process of the TFT architecture took 4732 minutes to converge to  the following set of hyperparameter values: $l$: 168, \# LSTM layers: 2, \# attention heads: 4, dropout: 0, hidden dimension size: 128, batch size: 1280. Given these values, each model of the ensemble was trained for 64 minutes on average, running for a total 46 epochs. The method required 1 minutes to produce forecasts for the complete test set.


\begin{table}[]
\centering
\caption{Evaluation of the selected DL methods on the test set (2021) in terms of forecasting accuracy (MAPE\% and RMSE) and computational cost (minutes). The first sub-column under each measure (MAPE\% and RMSE) lists its average value alongside the standard deviation of the 30 models involved in each ensemble. The second subcolumn lists the performances of the respective ensemble models (median of individual forecasts). The last column of the table reports the time required for tuning the hyperparameters of each architecture and training the models included in the ensembles.}\label{tab:2}
\begin{tabular}{@{}llllll@{}}
\toprule
\centering
\textbf{Architecture} & \multicolumn{2}{c}{\textbf{MAPE \%}} & \multicolumn{2}{c}{\textbf{RMSE}} &\textbf{Cost (minutes)} \\ 
~ & Individual models & Ensemble & Individual models & Ensemble & Tuning \& training \\
\midrule
 sNaive (24) & 6.52 & - & 551.54 & - & 1 \\
 sNaive (168) & 4.29 & - & 402.49 & - & 1 \\
 \hline
 MLP & 2.44 $\pm$ 0.15 & 2.02 & 211.62 $\pm$ 9.91 & 187.02 & 262 \\
 LSTM & 2.63 $\pm$ 0.14 & 2.23 & 259.1 $\pm$ 16.4 & 213.19 & 2,343\\
 N-BEATS & 2.43 $\pm$ 0.13 & 1.90 & 213.7 $\pm$ 13.2 & 181.68 & 2,983\\
 TCN & 2.52 $\pm$ 0.09 & 2.22 & 226.1 $\pm$ 5.8 & 205.25 & 5,349\\
 TFT & 3.36 $\pm$ 0.17 & 2.17 & 301.1 $\pm$ 14.6 & 208.13 & 6,653\\
\bottomrule
\end{tabular}
\end{table}

Taking into consideration the ensemble modeling approach (median of 30 networks) that followed the hyperparameter optimization process, the final performance of each architecture for year 2021 is summarized in Table \ref{tab:2}. Regarding the benchmarks, sNaive (128) performs much better than sNaive (24) as it replicates weekly seasonality which is significantly stronger in electricity load time series.

Regarding the DL methods, initially we observe that all DL architectures outperform the sNaive benchmarks. This is an expected result as sNaive models simply replicate past values and do not involve any learning process. Subsequently, we note that the feed-forward architectures, namely N-BEATS and MLP, clearly outperform the LSTM, TCN, and TFT models in terms of forecasting accuracy, with N-BEATS reaching the lowest MAPE of 1.90\% and RMSE of 181.68. The MLP follows with a significantly low error of 2.02\% and an also low RMSE of 187.02 despite the simplicity of its architecture and its limited computational requirements. This is an impressive result for an architecture that has not been originally designed for handling  sequential data. However it completely justifies its popularity and extensive usage along the years within time series forecasting and specifically STLF fields.

TCN and LSTM perform at a similar level in terms of MAPE (2.22\% and 2.23\% respectively), while the former scores better in terms of RMSE than both the latter and TFT --which exhibits a comparatively moderate MAPE of 2.17 despite the complexity of its architecture and its high computational requirements. Last but not least, it is notable that although the validation set mainly contained a period that has been highly affected by the COVID-19 pandemic \citep{Pelekis2022InPerformance}, the selected DL architectures manage to perform impressively well on average. Taking into consideration that the evaluation results derived by MAPE are in general alignment with those provided by RMSE, apart from insignificant deviations, from now on our study primarily focuses on MAPE to discuss further lower-level concepts regarding model accuracy.

\begin{figure*}[]
	\centering
        \includegraphics[width=0.65\textwidth]{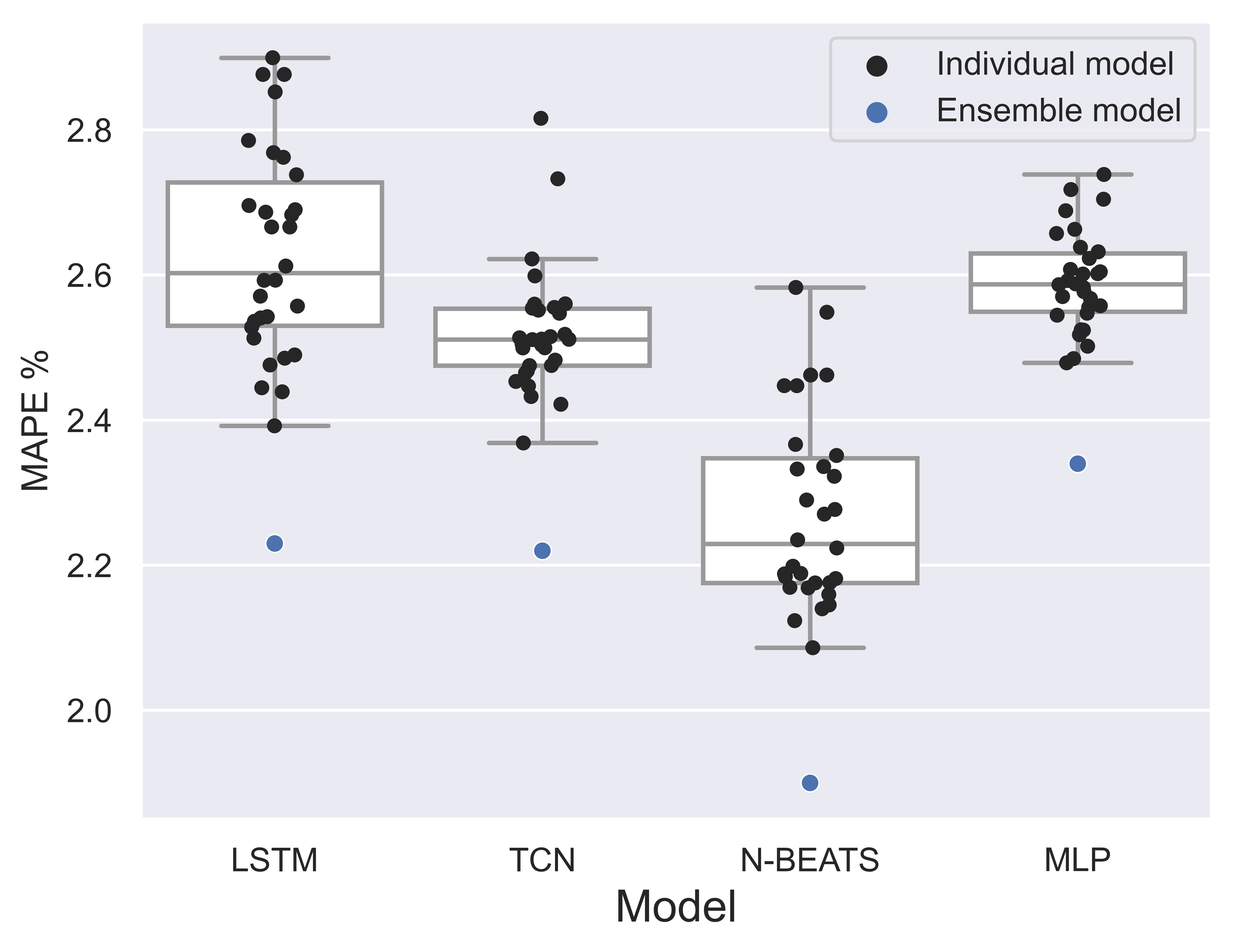}
	  \caption{Boxplots summarizing the accuracy (MAPE) of the 30 individual models (same hyperparameter values but different neural weight initializations) used to form the ensemble of each DL architecture. The MAPE of the ensemble is also depicted.}\label{fig:100}
\end{figure*}

To validate the added value of the ensembling strategy, Table \ref{tab:2} lists both the average forecast error of the individual models involved in each ensemble and the respective standard deviation of said errors. These results are also visualized in Figure \ref{fig:100}. As seen, the accuracy of the ensemble is always higher than that of the respective individual models, implying that the examined strategy cannot only ensure robustness by avoiding inaccurate forecasts, but also leads to superior accuracy overall. Note here the spectacular improvement (1.19\%) demonstrated by the TFT ensemble (2.17\%) compared to the respective individual models whose average error is (3.36\%). The latter can be attributed to the large variance among individual TFT forecasts errors that leads to significant neutralization of the median prediction. Note also that said large errors for the individual TFT models imply significant overfitting during the training process which is an obvious consequence of the large complexity of this model.

\begin{figure*}[]
	\centering
        \includegraphics[width=0.5\textwidth, angle=-90]{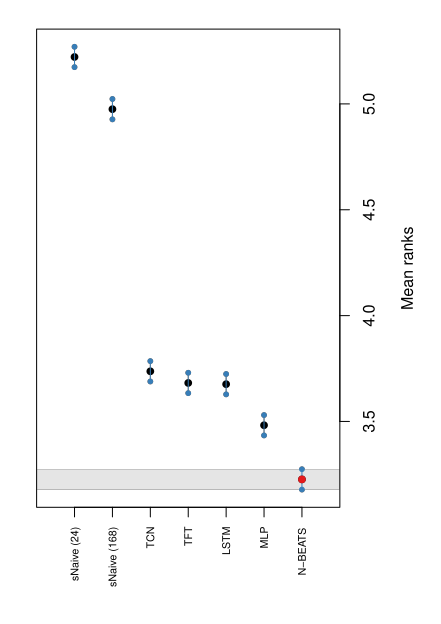}
	  \caption{Average ranks and 95\% confidence intervals of the four DL forecasting models and the two benchmarks considered in the present study. The multiple comparisons with the best test, as proposed by \citet{Koning2005TheResults}, is applied using MAPE for ranking the methods.}\label{fig:101}
\end{figure*}

To further investigate the differences reported between the examined DL methods and the benchmarks considered, we conduct the multiple comparisons with the best (MCB) test \citep{Koning2005TheResults}. The test computes the average ranks of the forecasting methods according to MAPE across the complete test set and concludes whether or not these are statistically different. Figure \ref{fig:101} presents the results of the analysis. If the intervals of two methods do not overlap, this indicates a statistically different performance. Thus, methods that do not overlap with the gray interval of the figure are considered significantly worse than the best, and vice-versa. The results of the MCB test confirm that N-BEATS and MLP are significantly more accurate than the rest of the methods with N-BEATS being the clear winner. The rest of the already discussed results are also confirmed by the MCB test.

\subsection{Explaining the performance of forecasting models}\label{sec:3.2}

\begin{figure*}[!htb]
\centering
\begin{subfigure}[b]{0.9\linewidth}
   \centering
   \includegraphics[width=\textwidth]{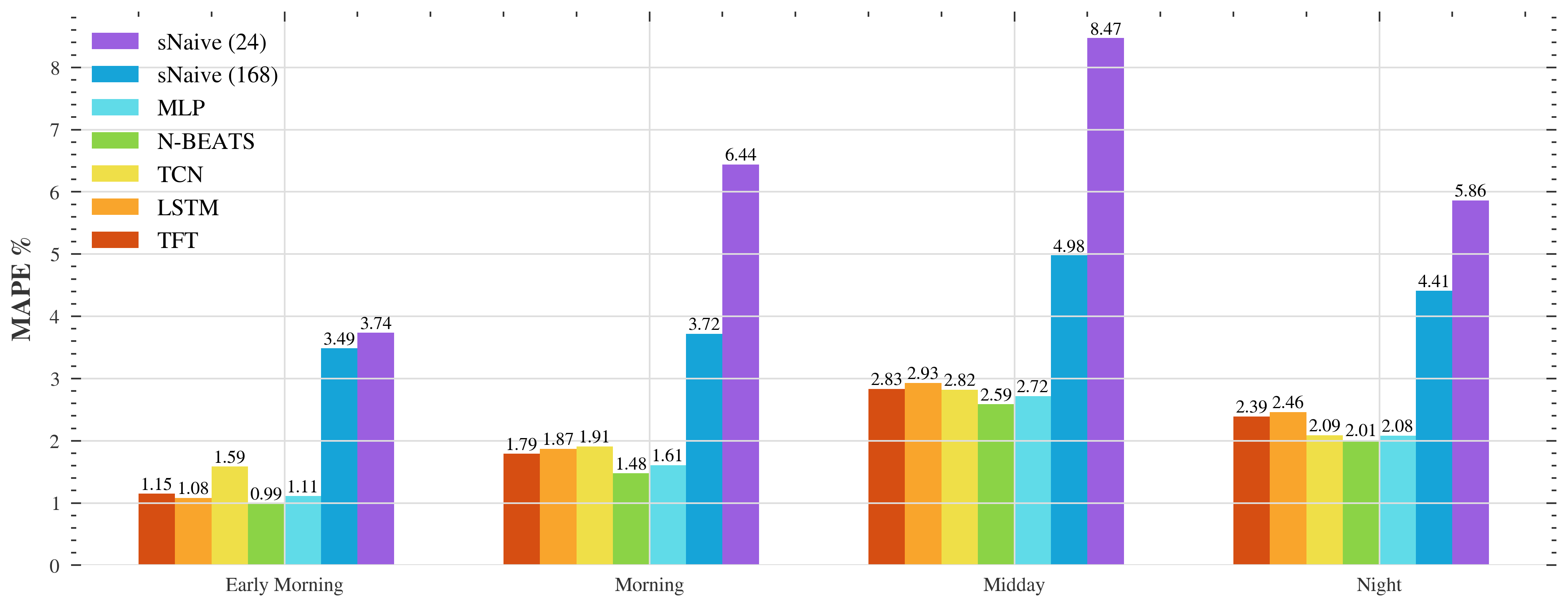}
   \caption{Performance of the models across different times of the day.}
   \label{fig:5.1}
\end{subfigure}
\hfill
\par\bigskip
\begin{subfigure}[b]{0.9\linewidth}
   \centering
   \includegraphics[width=\textwidth]{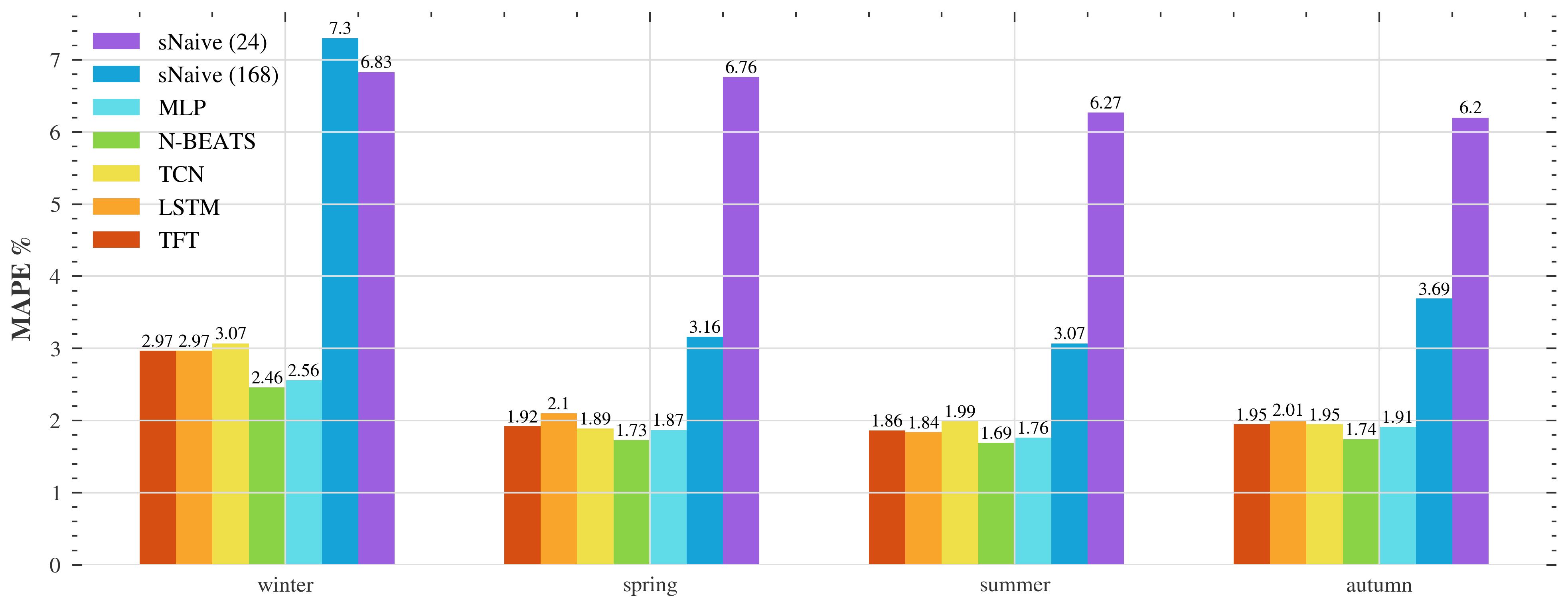}
   \caption{Performance of the models across different seasons.}
   \label{fig:5.2}
\end{subfigure}
\par\bigskip
\begin{subfigure}[b]{.45\linewidth}
   \includegraphics[width=\textwidth]{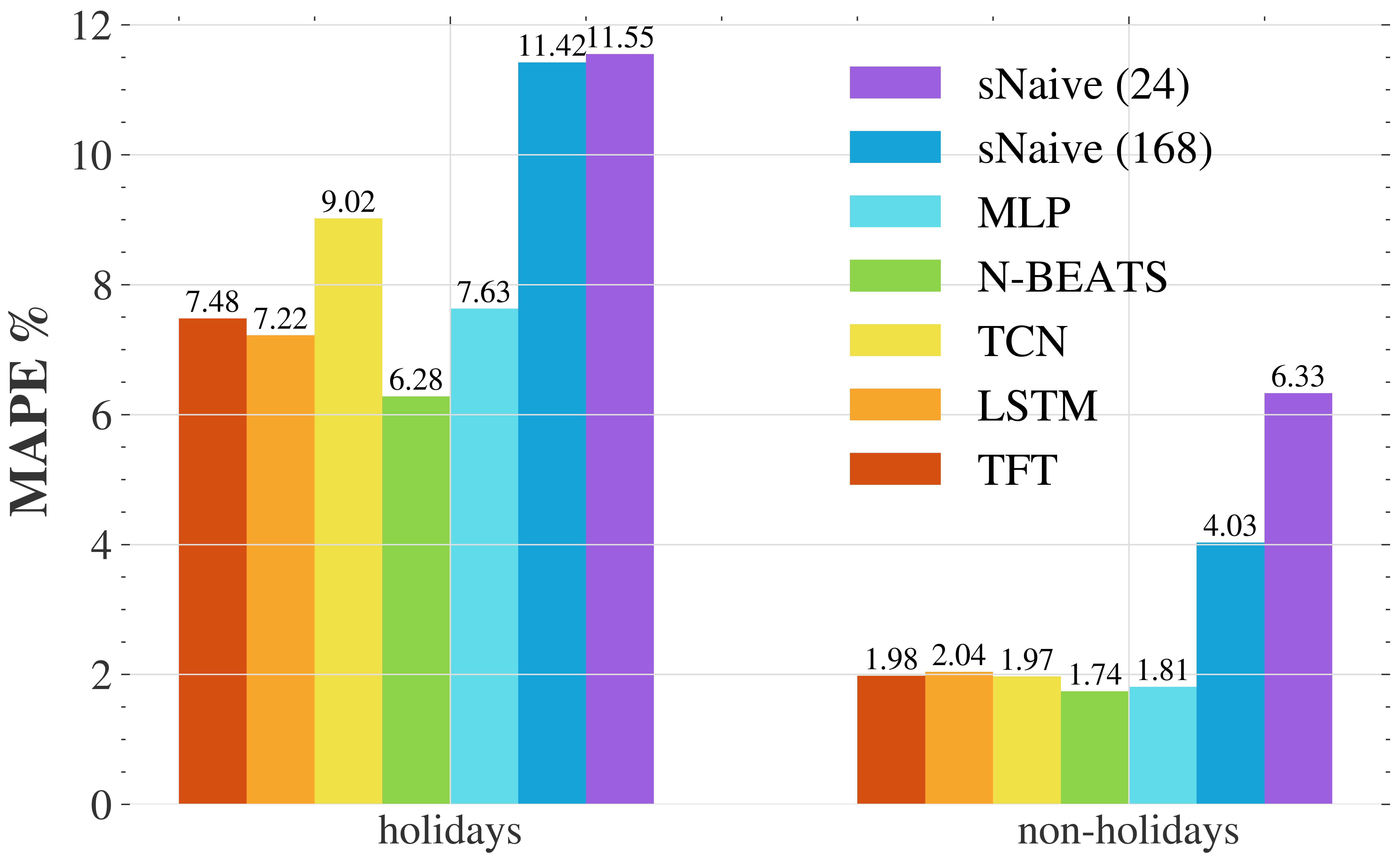}
   \caption{Performance of the models across holidays and non-holidays.}
   \label{fig:5.3}
\end{subfigure}
\hfill
\begin{subfigure}[b]{.45\linewidth}
   \includegraphics[width=\textwidth]{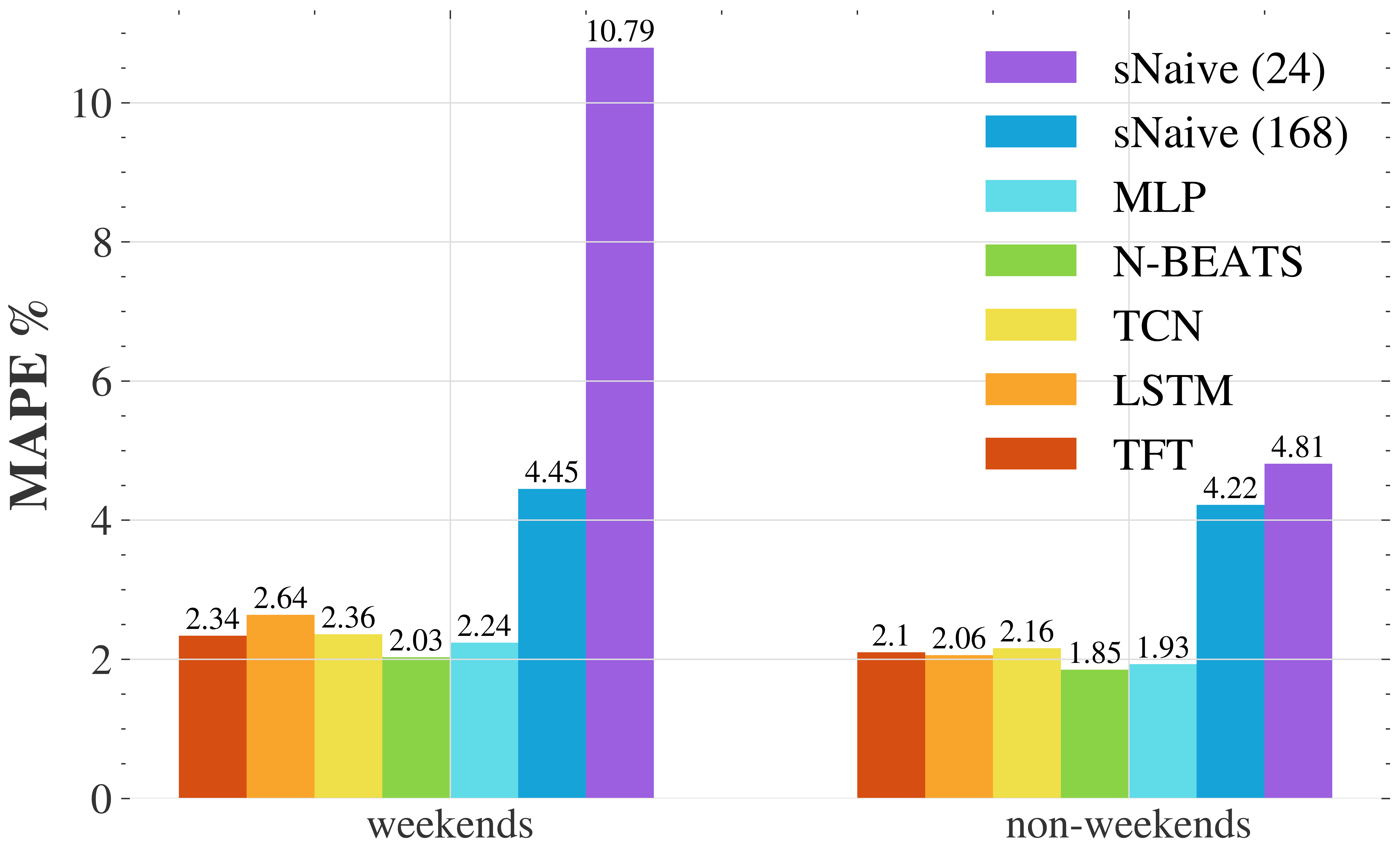}
   \caption{Performance of the models across weekends and non-weekends (weekdays).}
   \label{fig:5.4}
\end{subfigure}  
\caption{Performance of the models across different calendar features, as measured by MAPE.}\label{fig:5}
\end{figure*}

Table \ref{tab:3} presents the coefficients estimated by the MLR models, aiming to explain the forecasting performance achieved by the DL methods and the benchmarks using as accuracy drivers the selected calendar and weather features. The goodness-of-fit coefficient $R^2$ is also reported to measure the capability of the MLR models to identify strong correlations. Before interpreting the results of Table \ref{tab:3}, the following should be noted:

\begin{itemize}
    \item The calendar variables are binary and therefore their coefficients denote the extent to which the forecast error is increased on average during the time period they represent. On the contrary, temperature is a continuous variable, and the respective coefficients demonstrate the average increase in MAPE that is expected given a rise in the temperature by $1 ^\circ C $.
    
    \item The coefficients calculated by fitting the MLR models do not directly reflect the absolute fluctuations of the MAPE values across different models but rather the relative variation around the corresponding average value of a given model. This is a result of building each MLR model independently for each forecasting method. To address this issue, the complementary bar charts of Figure \ref{fig:5} are required to establish a complete comparison among the different models with regard to accuracy. The reader can also refer to appendix \ref{app:a} for additional bar charts that measure the forecasting accuracy using the RMSE metric. Note that the results of the RMSE are in general agreement to those produced by MAPE. However, they can provide useful insights on the variance of model performance across different time periods as we will explain later on.
    
    \item Drawing conclusions for the "early morning" hours and the "summer" periods is not possible due to the exclusion of these features from the MLR models to avoid severe multicollinearity problems. However, the selection of these variables was not random as these periods were usually characterized by lower forecast errors, thus exhibiting less interest for explanatory insights.
    
    \item  The considered features explain more than half of the MAPE variance ($R^2$ values higher than 0.5). Although including more external variables could result to higher $R^2$ values, given the randomness of the series and the challenges present in STLF, we suggest that this value still suggests an adequate amount of explainability. 
    
    \item All the coefficients of the MLR model have proved to be statistically significant with p-values lower than 0.05. This  reinforces the validity of the observations and conclusions extracted in the paragraphs to follow.
\end{itemize}


\begin{table}[h]
\centering
\caption{The estimated coefficients of the MLR models aiming to relate the MAPE values generated by sNaive (24), sNaive (168), MLP, LSTM, N-BEATS, TCN, and TFT in the test set of the study with the selected calendar and weather features.}\label{tab:3}
\begin{tabular}{@{\extracolsep{\fill}}l|lll|lll|l|l|l|ll}
\toprule
        ~ & \textbf{morning} & \textbf{midday} & \textbf{night} & \textbf{winter} & \textbf{spring} & \textbf{autumn} & \textbf{holiday} & \textbf{weekend} & \textbf{temperature} & \textbf{$R^2$} \\ \hline
        \textbf{sNaive (24)} & 2.96 & 4.65 & 2.26 & 1.30 & 0.93 & 0.38 & 4.92 & 6.04 & 0.06 & 0.58 \\
        \textbf{sNaive (168)} & 0.63 & 1.55 & 1.19 & 4.93 & 0.72 & 1.25 & 6.64 & 0.45 & 0.07 & 0.55 \\
        \textbf{MLP} & 0.60 & 1.56 & 1.05 & 1.08 & 0.33 & 0.38  & 4.49 & 0.34 & 0.02 & 0.57 \\
        \textbf{LSTM} & 0.93 & 1.93 & 1.56 & 1.28 & 0.44 & 0.41  & 3.80 & 0.62 & 0.01 & 0.57 \\
        \textbf{N-BEATS} & 0.63 & 1.62 & 1.15 & 1.04 & 0.29 & 0.26 & 3.58 & 0.21 & 0.02 & 0.54 \\
        \textbf{TCN} & 0.47 & 1.10 & 0.60 & 1.56 & 0.27 & 0.31 & 5.47 & 0.22 & 0.04 & 0.59 \\
        \textbf{TFT} & 0.85 & 1.81 & 1.48 & 1.37 & 0.35 & 0.40 & 4.18 & 0.31 & 0.01 & 0.57 \\
    \end{tabular}
\end{table}

\paragraph{\textbf{Benchmarks}}
As seen from Table \ref{tab:3} and Figure \ref{fig:5}, the forecast error of the sNaive (24) benchmark is not particularly affected by the season of the year as this factor cannot contribute much to the overall accuracy of the method given the short-term (daily) periodicity it assumes. However, sNaive (24) is significantly less accurate on holidays and weekends, which is expected given that said days have different load patterns than weekdays. Mediocre results are also observed for the "time of day" variables as switching among time periods affects the performance of sNaive (24) when transitioning from weekends to weekdays, and vice-versa. 

Regarding sNaive (168), the method is less sensitive to the "time of day" features (morning, midday, night) as by using the values of the previous week as forecasts the benchmark manages to capture the weekend to weekday transitions. Similar conclusions are true for the "weekend" variable as well. Nonetheless, on holidays, sNaive (168) still exhibits great errors given that holidays are not subject to weekly seasonality. Additionally, as holidays tend to be concentrated within small periods of time (e.g. Christmas period)--also accompanied by other semi-vacation days due to days off that are given by most companies to their employees-- forecasts of weekly seasonality result to low accuracy for long periods of time, both including vacation days and the first week to come after them. With respect to the "season" variables, the forecast error of the sNaive (168) method is negatively affected in winter, which can be attributed to i) the aforementioned irregularity within the Christmas holidays period and ii) the fact that changes in temperature among different weeks can lead to significant load curve alterations during the winter, given Figure \ref{fig:x}. Similarly, "autumn" exhibits a relatively large coefficient value, potentially caused by the switch from summer to autumn that signals the return from holidays and thus an expected notable variation in the load curve from one week to another. 

It can be also observed that the increase of the temperature reduces the accuracy of both benchmarks, demonstrating a certain increase in the variability of the time series that cannot be captured effectively by the naive models.  

\begin{figure*}[!htb]
	\centering
        \includegraphics[width=0.6\linewidth]{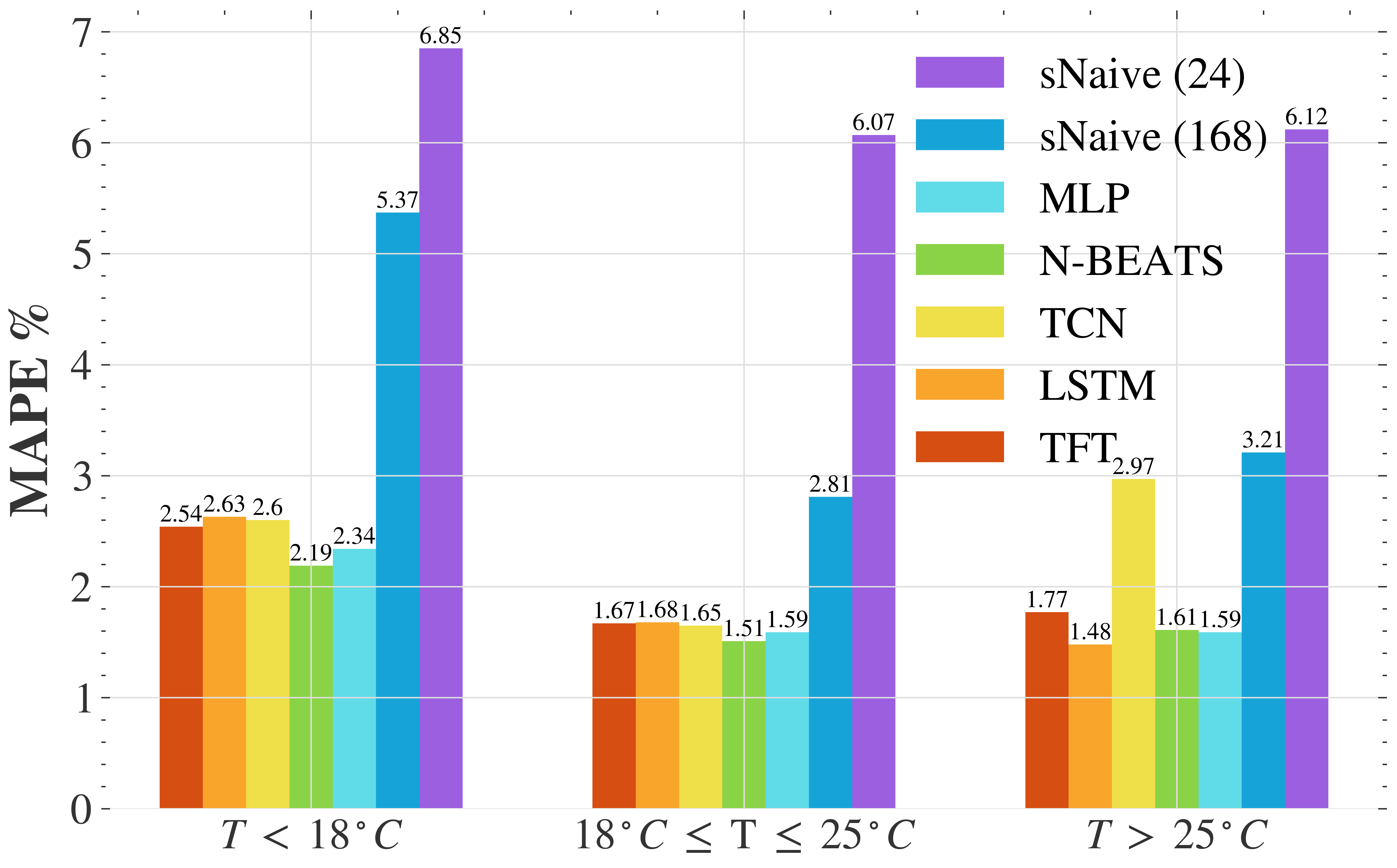}
	  \caption{Performance of the models across average daily temperature levels}\label{fig:6}
\end{figure*}

\paragraph{\textbf{DL models}} 
Before discussing the MLR coefficients of the DL models and their absolute performance across the the different values of external variables based on the bar plots of Figure \ref{fig:5}, it should be first noted that all DL architectures perform better overall compared to the naive benchmarks, irrespective of the feature in question. This is expected and confirmed by Figure \ref{fig:5}. This observation is not to be confused with the magnitude of the MLR coefficients as naive models often involve smaller values than DL models, thus denoting a smaller relative increase to their average forecasting error, caused by a specific external variable. An additional preliminary observation is that, in general, all models mostly demonstrate a certain performance consistency meaning that they do not deviate significantly in terms of ranking across the studied calendar and weather features. In the same context, the feed-forward networks i.e. N-BEATS and MLP demonstrate a consistent superiority against the rest of the architectures with nearly no exceptions at all to that for N-BEATS. Therefore, in the rest of this section we mainly discuss what noticeably deviates from this basis.  

Starting from the "time of day" variables, according to the MLR model of Table \ref{tab:3}, MLP, LSTM, N-BEATS, and TFT seem to be affected in a similar fashion to sNaive (168) by the alteration of seasons, exhibiting relatively low coefficient values that indicate a satisfactory modeling of the daily seasonal pattern. Specifically during midday and night, the high variability and the frequent changes in the monotony of the load and the peak load (see Figure \ref{fig:1}), lead to larger forecast errors compared to other calendar features, with TCN being however less affected by said changes compared to the rest of the DL models already implying a somehow underfitted model. In general, N-BEATS and MLP demonstrate the best performances from a "time of day" perspective while the rest of the models exhibit alternating behavior. However, while N-BEATS tends to perform better in general, the MLP produces also notably accurate forecasts during the night which is a period of peak load and significant variability. Note here that midday is the less forecastable time interval for all models due to its volatility across seasons, and especially cold and hot days, both in terms of load magnitude and shape. Also note that higher accuracy in terms of MAPE during hours of peak load, such as night hours, can usually lead to inconsistencies between MAPE and RMSE rankings as the presence of high load values during this interval results to higher absolute errors (RMSE) given a specific relative error (MAPE). In this context, this is the period where TCN: i) highly outperforms its close competitors (TFT, LSTM) and as from \ref{fig:3.1}, and ii) its performance is close to that of the best models (N-BEATS and LSTM). Hence, we can confirm that the former deals better with peak load hours. Similarly, a slight alteration can be also observed in the ranking between N-BEATS ans MLP according to the RMSE, implying the more accurate forecasting by MLP of the peak loads observed in this interval.


With respect to calendar variables relating to the seasons of the year, the DL models exhibit a certain insensitivity to the respective features (especially spring and autumn). This is confirmed by the relatively small coefficient values of Table \ref{tab:3} alongside Figure \ref{fig:5.2} where it is found that all DL models produce errors that are relatively close to their average. It can be observed, however, that during winter months the forecastability of the models decreases significantly compared to other seasons, leading to both increased MLR coefficients and average MAPE values. This behavior can be justified by Figure \ref{fig:1.3} where it was observed that the average daily load profile during the winter period significantly differs from the rest of the seasons both in terms of shape and magnitude due to the consistent electricity demand by heating systems. Note here that among the DL models, N-BEATS and MLP seem to perform better during this period, capturing the winter load patterns more sufficiently. Similarly to the night variable, the TCN demonstrates lower accuracy in terms of MAPE compared to TFT and LSTM but exhibits a significantly higher accuracy in terms of RMSE which again implies its good performance on peak load forecasts that are even higher during the winter. Similar conclusions can be drawn again for the MLP. Therefore, it is confirmed that MLP and TCN tend to focus on peak load hours compared to their close competitors (N-BEATS and TFT, LSTM respectively). This can be attributed to the fact that such networks, given their relative implementation simplicity, tend to stick to their training objective (L2 loss) ignoring the reduced load periods that have low impact on it.

With regard to holidays, TCN seems to be more sensitive than sNaive (24) as it can be observed from the notably larger MLR coefficients of said models (see Table \ref{tab:3}). Regarding the absolute performance, all DL models perform better on average than the benchmarks. In general, holidays seem to be the feature that most deteriorates forecast accuracy. This becomes obvious in Figure \ref{fig:1.3} where a large increase from the average MAPE is observed for all DL models. In this case, LSTM and particularly N-BEATS exhibit the lowest deficits, reaching a MAPE of 7.22 \% and 6.28 \%, respectively. The relatively small error of the latter model demonstrates the ability of N-BEATS to capture the substantial alterations of the load curve during national holidays, confirming its robustness and superiority throughout distribution shifts that tend to appear less frequently within an electricity load time series. Again, based on the previous observations on MLP and TCN regarding underfitting and limited focus on low load periods the low accuracy here is expected. With respect to TFT, it seems to preserve its moderate performance profile similar to most of the previous cases. This fact that can be attributed to the large variance of the individual models forming the TFT ensemble, therefore leading to a certain neutralization of any extreme behaviors.

Concerning the weekend variable, all models seem to be less sensitive than both naive methods given their MLR coefficients (see Table \ref{tab:3}), apart from the LSTM. This can be credited to the vanishing gradient problem of LSTM networks as they always need to propagate the gradient for at least 5 days in the past during training in order to access the required knowledge for weekend forecasts. Nevertheless, since weekends are observed with a certain periodicity  and the input vector ($l$) of all DL models is larger than a week, weekends are relatively forecastable, resulting to low MLR coefficient values. In this case, TCN and N-BEATS have the lowest coefficient, while TFT, and MLP, fall somewhere in the middle. Regarding the absolute performance of models (Figure \ref{fig:5}), with the exception of the poor performing LSTM, it can be confirmed that both weekdays and weekends are forecast with near to average errors, however with a relatively reduced performance on weekends instead of weekdays. The latter can be attributed to the fact that weekend days are fewer in number while they are mostly related with low load levels, therefore leading to insufficient optimization of the L2 loss during model training, as already discussed.

With respect to the temperature, according to the MLR model, this variable also has a notable effect on forecast accuracy. This is because, although the estimated coefficients are of relatively low values, the temperature variable is the only non-binary one, thus implying the multiplication of said coefficients with values that range from -1 $^{\circ}$C to 36.4 $^{\circ}$C and therefore a significant impact on the forecast error. This finding is also evident in Figure \ref{fig:6}, where we find that lower temperatures typically result to higher forecast errors. This is expected as, during the winter, slight changes in the temperature can significantly alter the load magnitudes rendering the period harder to predict (see Figure \ref{fig:x}) as above mentioned. An exception to that is the TCN, which, during hot days, exhibits impressively large absolute errors, which can be attributed to the distortion of the respective load curve and the missing peak during the night which has been the main interval of excellence for this model. Nonetheless, such conclusions cannot be considered safe taking into consideration the high multicollinearity of the temperature with the season and time of day factors alongside the multimodal nature of the joint load and temperature distribution (see Figure \ref{fig:x}). Therefore, the following section focuses on the explicit effect of the temperature on the forecast error by investigating its value as a predictor variable within the examined DL models.

\section{Discussion}\label{sec:4}

Motivated by the exploratory data analysis of Section \ref{sec:2.1} and the latter insufficiency of the MLR model (see Section \ref{sec:3.2}) in clearly explaining the effect of the temperature variable on the forecast errors of the models, in this section we explicitly investigate the contribution of the temperature explanatory variable to the forecast error. In this direction, we include the temperature as an input variable of the already implemented DL models (LSTM, MLP, N-BEATS, TCN, TFT), by extending the univariate setup that has been the focus of this study until now. Our analysis specifically focuses on examining the correlation between load forecasts using historical temperature inputs (referred to as "ex-post forecasts") rather than relying on historical forecast inputs (known as "ex-ante forecasts"). We do that for two main reasons: i) ex-ante forecasts are not publicly available and especially for that far in the past (from 2013 and on which is the beginning of our training data set), ii) much of the literature makes use of ex-post forecasts \cite{Haben2019a}, aiming to investigate the impact of a hypothetically ideal weather forecast on model performance. Note here though, that in the context of real-world STLF use cases, the ex-post setup is not realistic as real temperature values are not a priori known and therefore the acquisition of ex-ante forecasts is required. This implies the need for access to (usually paid) application programming interfaces (APIs) and databases, both for real-time and historical forecasts. Subsequently, such data ingestion into production DL models, requires advanced model development skills, which are not always the case for EPES and STLF stakeholders.

The model training methodology for the multivariate load and temperature setup for the five DL models of interest follows similar steps with the above described univariate problem. In this context, the best setups from the univariate optimization process of Section \ref{sec:3.1} are employed again using the ensembling process described in Section \ref{sec:2.2}. However, this time the ensembles are trained in a multivariate setting that also incorporates the ex-post forecasts of the temperature in Lisbon. The results of said multivariate setup are summarized in Table \ref{tab:ex-post}.

\begin{table}[]
\centering
\caption{Comparison of the results of the selected DL ensemble models on the test set (2021) in terms of forecasting accuracy (MAPE\%). The first column refers to the results of the univariate versions of the models that have been also listed in Table \ref{tab:2}. The second column refers to the performance of the multivariate setup that also incorporates the ex-post forecasts of the temperature in Lisbon as a predictor variable.}\label{tab:ex-post}
\begin{tabular}{@{}lll@{}}
\toprule
\centering
\textbf{Architecture} & \multicolumn{2}{c}{\textbf{MAPE \%}} \\ 
~ & Univariate & Multivariate (ex-post) \\
\midrule
MLP & 2.02 & 1.92\\
LSTM & 2.23 & 2.20 \\
N-BEATS & 1.90 & 2.00 \\
TCN & 2.22 & 2.12\\
TFT & 2.17 & 2.06 \\
\bottomrule
\end{tabular}
\end{table}

From Table \ref{tab:ex-post} it can be observed that all models except for N-BEATS exhibit slightly better forecasting performance in the multivariate setup. This improvement is expected as, according to the analysis of Sections \ref{sec:2.1} and \ref{sec:3.2}, the temperature variable appeared to significantly affect the daily load curves and forecast errors (even though the temperature is reported for a single weather station in the city of Lisbon). Hence, our expectations are generally confirmed, except for the case of N-BEATS. This behavior can be attributed to the fact that N-BEATS has been originally designed for handling univariate forecasting tasks, as well as to its sophisticated architecture that requires the implementation of both forecasting and backcasting processes, rendering the learning process more challenging in multivariate settings. Similar conclusions for the interaction of N-BEATS with external variables have also been drawn by \citet{Pelekis2022InPerformance}, therefore validating our findings. Nonetheless, what is notable here is that, despite the improvement of the rest of models, none of them manages to outperform the initial univariate implementation of N-BEATS which still can be considered the "winner" of our comparative study. As a final note, it should be mentioned that since the improvements of the multivariate models are based on ex-post forecasts, which would be difficult to estimate precisely in real-life conditions, the actual differences between the multivariate and univariate variants of the examined DL models could be less significant in practice.

\section{Conclusions}\label{sec:5}

This study conducted a comparative analysis of state-of-the-art time series forecasting DL models within the STLF field, focusing on producing univariate day-ahead forecasts for the Portuguese national net aggregated electricity load. The MLP, LSTM, N-BEATS, TCN, and TFT architectures were examined, following a proper hyperparameter tuning and ensembling process.

Our results suggest that feed-forward architectures are still more than capable of producing state-of-the-art forecasts despite the recent predominance of recurrent neural setups in the domain. In particular, N-BEATS consistently outperforms the rest of the DL models in terms of MAPE, reaching a value of 1.90\%, while the MLP comes right after it with a MAPE of 2.02 \%. TFT follows with 2.17\% despite its high computational requirements and complexity, while TCN and LSTM closely follow with similar performance (2.23\% and 2.22\% respectively). As a result, the relevance of residual stacked feed-forward NNs is STLF confirmed.

Considering the role of accurate STLF within the power grid, the reported improvements in the forecast error can lead to a significant reduction of annual operation costs for system operators and utilities due to more effective handling and management of generation planning (spinning reserves, outage reduction), imbalance markets, and demand side management. Specifically, compared to MLP, N-BEATS leads to an improvement in MAPE of 0.12\%. According to \citet{StimmelCarol2019BigGrid}, if such an improvement can be also extended to more granular energy forecasts, at least \$3 million can be saved in operating costs of imbalance markets.

Our results also highlight that ensembling can effectively mitigate the uncertainty of neural weights initialization, resulting to significantly more accurate forecasts than any individual model. Note that amongst the two top performing models, although N-BEATS performs significantly better, the MLP requires significantly lower computational resources (approximately 11 times), which may be an important factor to consider for stakeholders that possess limited computational resources, seek more energy efficient and sustainable machine learning operations and have to re-train their models at a regular basis.

In order to investigate the factors that drive the performance of each univariate DL model, a MLR model was built per case, correlating the MAPE values of the model with a selected set of key calendar and weather features. In this context, national holidays, cold/winter days and midday hours, proved to be the least forecastable instances overall. The results of this analysis can serve as guidelines for EPES stakeholders for forecasting model selection based on their daily forecast needs across days, seasons, and temperature patterns, always taking into consideration that such industrial stakeholders often require well-established architectures with publicly available implementations rather than the developing new DL architectures from the ground up. In this direction, they can benefit from being able to select and configure their forecasting models on-demand, hence maximizing the added value that can be offered by each one of them depending on the above mentioned external variables. The most important of these guidelines include:

\begin{itemize}
\item the option of N-BEATS for national STLF tasks as an all-around solution with good overall performance;
\item the option for MLP instead of N-BEATS for a well-balanced trade-off between forecast accuracy and computational cost; 
\item the option of MLP instead of N-BEATS in case the stakeholder is exclusively interested in forecasting peak load periods;  
\item the avoidance of sophisticated models such as TFT due to their high complexity and computational requirements;
\item the selection of TCN instead of TFT or LSTM in case the stakeholder is exclusively interested in forecasting peak load periods; 
\item the avoidance of LSTM for weekend day forecasts due to the unwanted effects of the vanishing gradient problem on the involved long-term dependencies.
\end{itemize}

As a final experimentation step, we explicitly investigated the impact of the temperature in Lisbon to the national forecast error by incorporating it as a predictor in our forecasting schemes. Our results indicate that ex-post forecasts can contribute to the improvement of load forecasts for all models with the exception of N-BEATS, which univariate implementation is still however more accurate than any multivariate approach. In real-world use cases this is based on the assumption of the availability of highly reliable historical and real-time weather forecasts which come at cost of data purchases and high programming skills.

\section{Future work}\label{sec:6}
Regarding future perspectives, this comparative assessment could be replicated using data from more European countries, aiming to benchmark the performance of state-of-the-art DL models and investigate the impact of key accuracy drivers at a larger scale. Additionally, the set of the examined DL architectures could be further expanded to include even more models that have been recently proposed in the field. A fully automated MLOps framework is also envisaged, with the objective to enable an automated, off-the-shelf toolkit for model training, validation, evaluation, ensembling and deployment for EPES/ STLF stakeholders, such as transmission system operators, distribution system operators, aggregators, and energy market operators. From a model selection perspective, the creation of a hybrid model could be examined, one that would automatically perform model selection conditioned by the external calendar features and temperatures forecasts, based on the periods of excellence of its contained baseline models. As an additional future perspective of the current study, and similar to \citet{Hobbs1999AnalysisForecasts, Ortega-Vazquez2006EconomicInterruptions, StimmelCarol2019BigGrid}, an attempt for an accurate and up-to-date quantification of the impact of national load forecasting errors on grid planning, operation and utility revenues would be valuable piece of research. Finally, a future additional training procedure of the models with the inclusion of ex-ante forecasts for Lisbon or even additional regions would be crucial to approximate the actual benefit of multivariate STLF.

\section*{Acknowledgment}
This work has been funded by the European Union’s Horizon 2020 research and innovation programme under the I-NERGY project, grant agreement No. 101016508. Additionally, the HPC resources utilized for training and optimizing the required machine learning models in this study have been provided by the EGI-ACE project, which also receives funding from the European Union’s Horizon 2020 research and innovation programme under grant agreement No. 101017567.









\bibliographystyle{cas-model2-names}

\bibliography{references}


\onecolumn
\appendix
\section{Figures} \label{app:a}
\renewcommand\thefigure{\thesection.\arabic{figure}}
\setcounter{figure}{0} 
\begin{figure*}[!htb]
\centering
\begin{subfigure}[b]{0.9\linewidth}
   \centering
   \includegraphics[width=\textwidth]{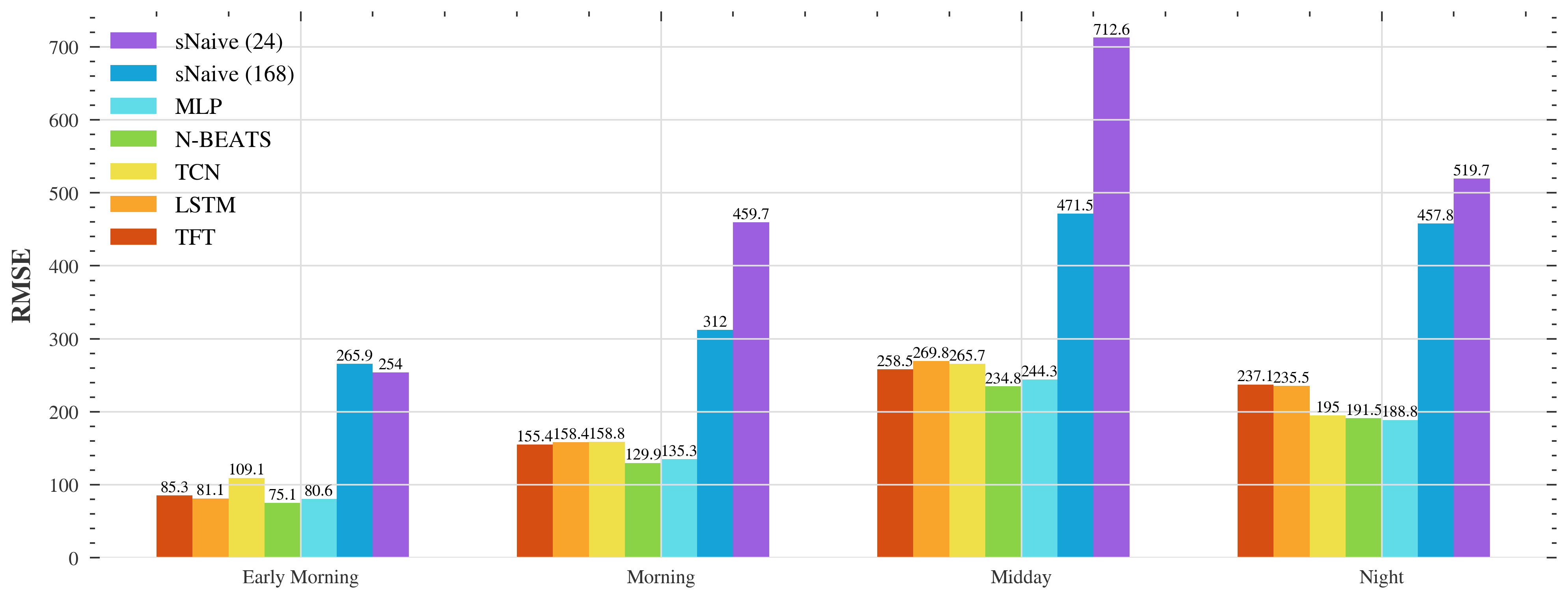}
   \caption{Performance of the models across the times of day.}
   \label{fig:3.1}
\end{subfigure}
\hfill
\par\bigskip
\begin{subfigure}[b]{0.9\linewidth}
   \centering
   \includegraphics[width=\textwidth]{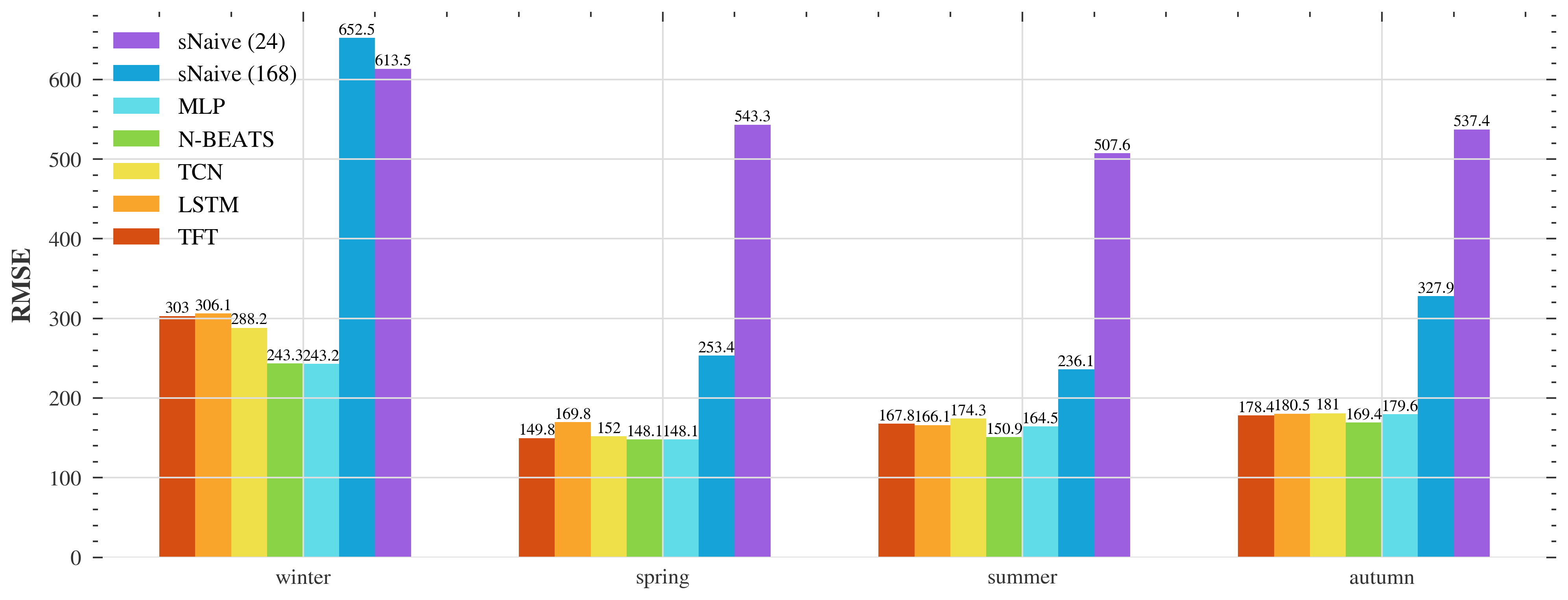}
   \caption{Performance of the models across seasons.}
   \label{fig:3.2}
\end{subfigure}
\par\bigskip
\begin{subfigure}[b]{.45\linewidth}
   \includegraphics[width=\textwidth]{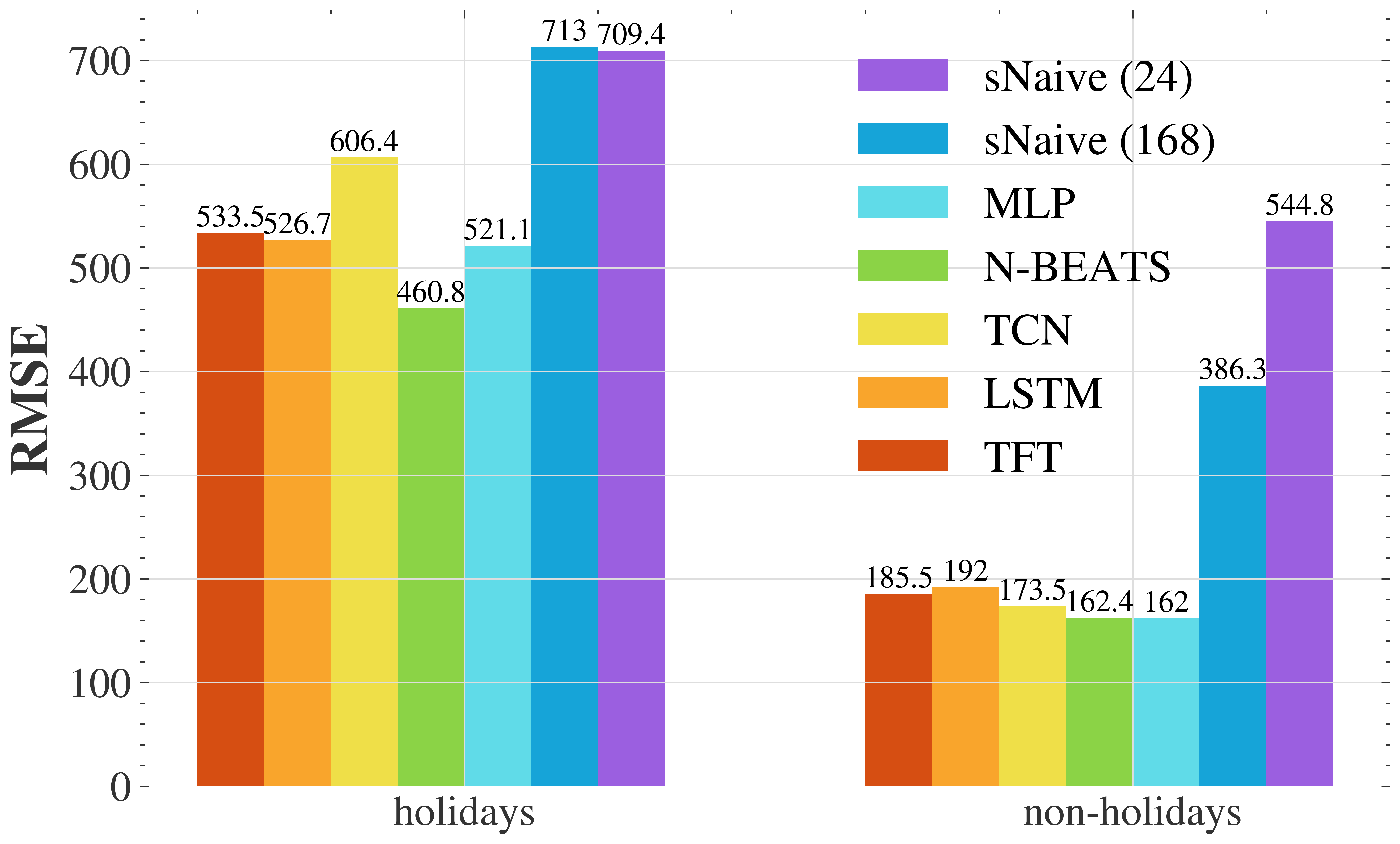}
   \caption{Performance of the models across holidays and non-holidays.}
   \label{fig:3.3}
\end{subfigure}
\hfill
\begin{subfigure}[b]{.45\linewidth}
   \includegraphics[width=\textwidth]{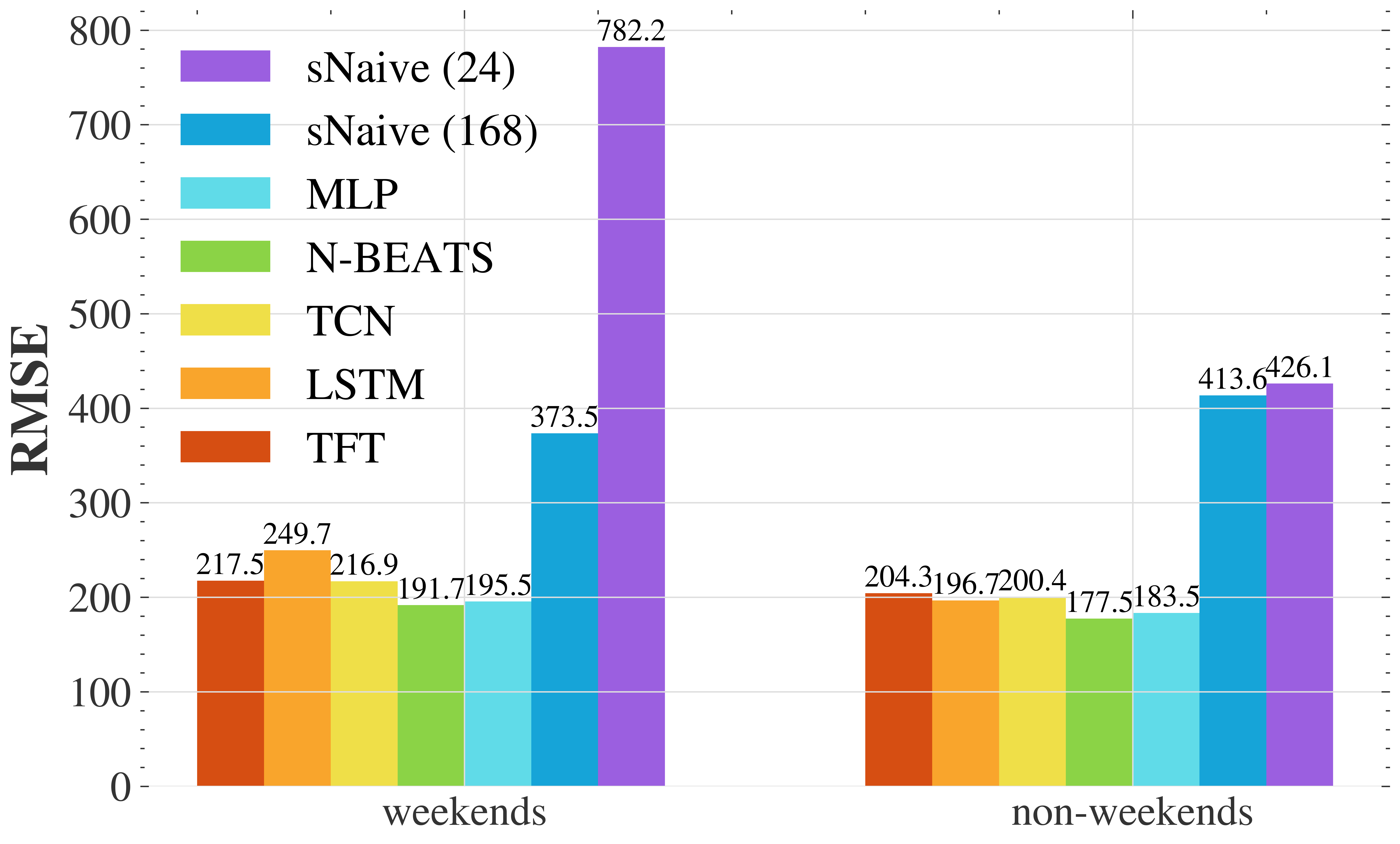}
   \caption{Performance of the models across weekends and non-weekends (weekdays).}
   \label{fig:3.4}
\end{subfigure}  
\caption{Performance of the models across external variables measured by the RMSE evaluation metric}\label{fig:3}
\end{figure*}

\begin{figure*}[!htb]
	\centering
        \includegraphics[width=0.6\linewidth]{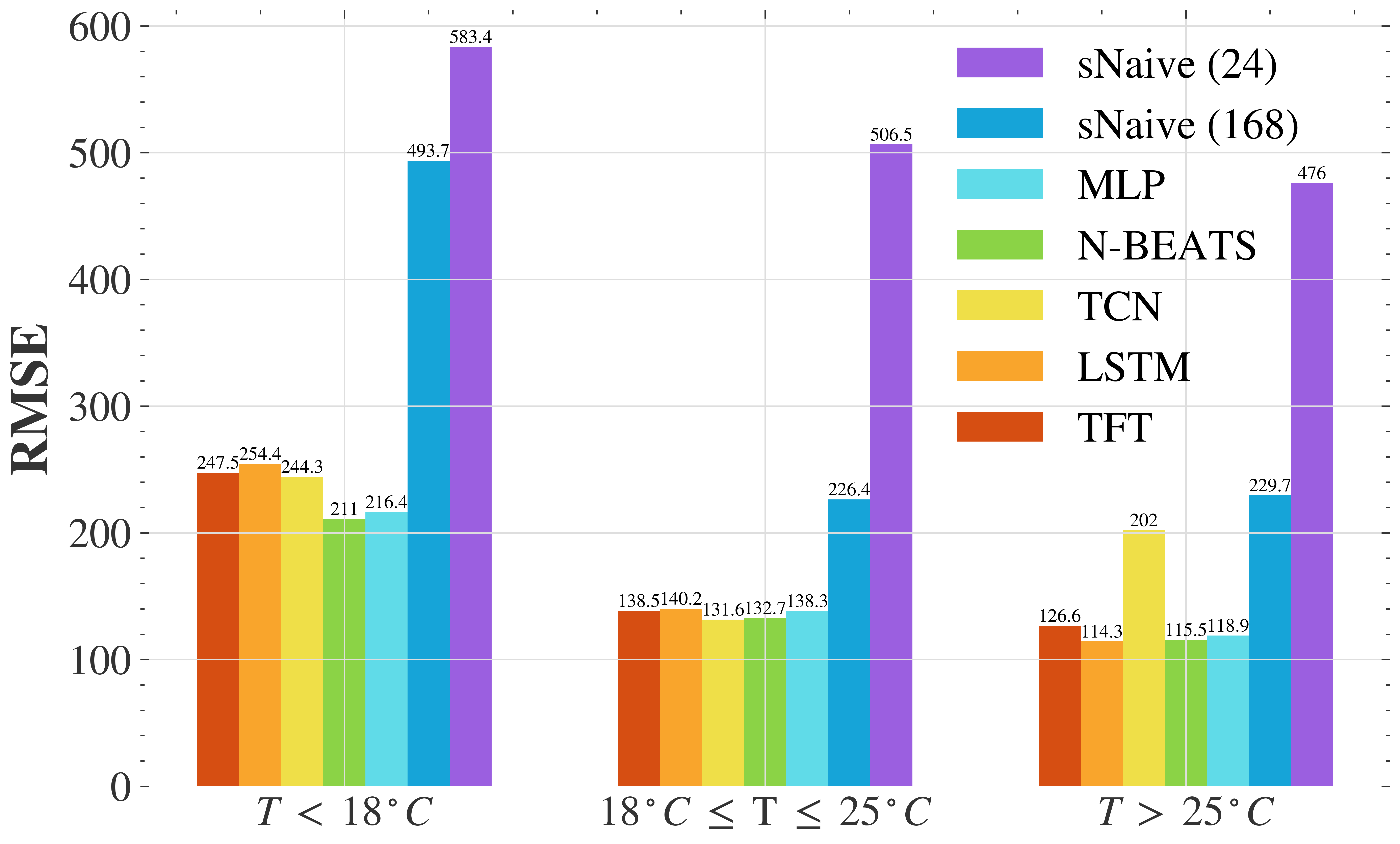}
	  \caption{Performance of the models across average daily temperature levels}\label{fig:barplot_temp_rmse}
\end{figure*}

\end{document}